\documentclass{article}

\usepackage[final]{corl_2026} 

\usepackage{wrapfig}
\usepackage{algorithm}
\usepackage{algorithmic}

\usepackage{lipsum}
\usepackage{amsmath}
\usepackage{amssymb}
\usepackage{mathtools}
\usepackage{amsthm}
\usepackage{graphicx}
\usepackage{amsmath, amssymb, amsthm}
\newtheorem{theorem}{Theorem}
\usepackage{marvosym}

\usepackage{dsfont} 
\usepackage{booktabs} 
\usepackage{tabularx}   
\usepackage{multirow}   
\usepackage{ragged2e}   
\usepackage{array} 
\usepackage{makecell}             
\usepackage{caption, subcaption, overpic, textpos}

\title{RoboMemory: A Brain-inspired Multi-memory Agentic Framework for Interactive Environmental Learning in Physical Embodied Systems}

\author{
    \textbf{Mingcong Lei}\textsuperscript{1,3}\textsuperscript{*} \textbf{,} 
    \textbf{Honghao Cai}\textsuperscript{3}\textsuperscript{*} \textbf{,} 
    \textbf{Yuyuan Yang}\textsuperscript{3}\textsuperscript{*} \textbf{,}
    \textbf{Yimou Wu}\textsuperscript{6}\textsuperscript{*} \textbf{,} 
    \textbf{Jinke Ren}\textsuperscript{2,1} \\
    \textbf{Zezhou Cui}\textsuperscript{1,3}\textbf{,} 
    \textbf{Liangchen Tan}\textsuperscript{4}\textbf{,} 
    \textbf{Junkun Hong}\textsuperscript{1}\textbf{,}
    \textbf{Gehan Hu}\textsuperscript{1,3}\textbf{,} 
    \textbf{Shuangyu Zhu}\textsuperscript{1} \\ 
    \textbf{Shaohan Jiang}\textsuperscript{1,3}\textbf{,} 
    \textbf{Ge Wang}\textsuperscript{1,3}\textbf{,} 
    \textbf{Junyuan Tan}\textsuperscript{1}\textbf{,} 
    \textbf{Zhenglin Wan}\textsuperscript{5}\textbf{,} 
    \textbf{Zhongkai Xue}\textsuperscript{3} \\
    \textbf{Zheng Li}\textsuperscript{6}\textbf{,} 
    \textbf{Zhen Li} \textsuperscript{2,1 \Letter} \textbf{,}
    \textbf{Shuguang Cui}\textsuperscript{2,1} \textbf{,}
    \textbf{Yiming Zhao}\textsuperscript{7} \textbf{,}
    \textbf{Yatong Han}\textsuperscript{7 \Letter}    \\
    \textsuperscript{\rm 1} Shenzhen Future Network of Intelligence Institute (FNii-Shenzhen) \\
    \textsuperscript{\rm 2} School of Science and Engineering, The Chinese University of Hong Kong, Shenzhen \\
    \textsuperscript{\rm 3} The Chinese University of Hong Kong, Shenzhen 
    \textsuperscript{\rm 4} The University of Hong Kong \\
    \textsuperscript{\rm 5} National University of Singapore
    \textsuperscript{\rm 6} The Chinese University of Hong Kong
    \textsuperscript{\rm 7} Ising AI \\
    \texttt{lizhen@cuhk.edu.cn}\quad\texttt{hanyatong@cuhk.edu.cn} \\ 
    Project website: \href{https://sp4595.github.io/robomemory/}{https://sp4595.github.io/robomemory/} \\
}

\begin{document}

\onecolumn{
\renewcommand\onecolumn[1][]{#1}
\maketitle
\begin{center}
    \label{fig:intro}
    \captionsetup{type=figure}
    \includegraphics[width=0.95\textwidth]{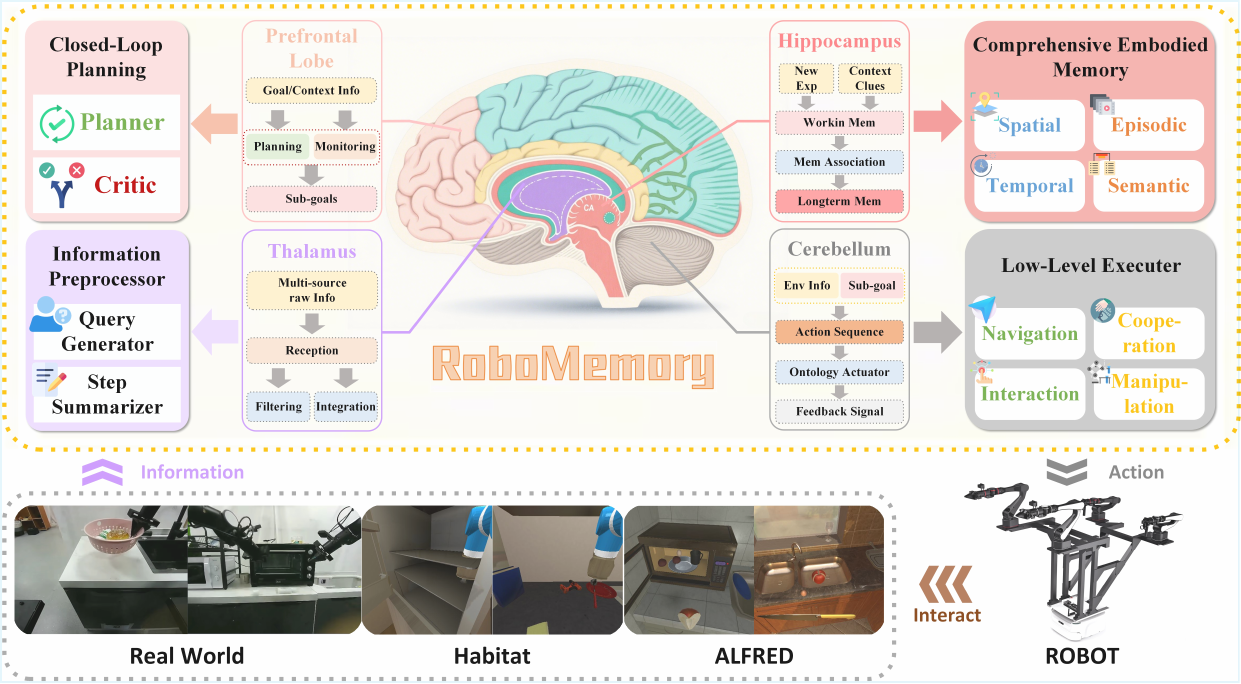}
    \captionof{figure}{RoboMemory adopts a brain-inspired organization of complementary memory and control modules, enabling long-term planning and experience reuse across diverse environments (real-world, Habitat, ALFRED) and robotic hardware.}
\end{center}
}


\begin{abstract}
    Long-horizon embodied tasks require agents to learn, reason, and generalize robustly across complex real-world environments. However, existing approaches often struggle with partial observability, fragmented spatial reasoning, and inefficient integration of heterogeneous memories, limiting their capacity for long-horizon adaptation. To address this, we introduce RoboMemory, a brain-inspired framework that unifies Temporal, Spatial, Semantic, and Episodic memory for efficient long-horizon planning and interactive learning. Its core innovations are a comprehensive memory system designed for embodied tasks, and a dynamic spatial knowledge graph for scalable, consistent memory updates. Extensive experiments on EmbodiedBench show that RoboMemory improves the average success rate of Qwen2.5-VL-72B-Ins by 26.5 percentage points and achieves performance comparable to strong closed-source VLM agents. Real-world trials further suggest RoboMemory's potential for interactive environmental learning by reusing accumulated experience across repeated task attempts.
    These results demonstrate RoboMemory as a scalable memory-augmented framework for embodied agents, improving long-horizon planning and cumulative learning in complex interactive environments.
\end{abstract}


\keywords{Task and motion planning, Memory, Embodied Agent} 


\section{Introduction}

The lack of persistent, structured memory remains a critical bottleneck for applying Vision-Language Models (VLMs) \citep{hurst2024gpt,bai2025qwen2} to long-horizon embodied tasks. Although recent post-training methods have improved the embodied planning and manipulation abilities of VLM agents \citep{team2025robobrain}, they still struggle in scenarios that require frequent navigation, object interaction, and long-term decision making. In such settings, an agent must operate under partial observability, handle dynamic occlusions, remember previously explored regions, and update its plan as the environment changes. Existing context-window-based agents often lack a persistent global spatial representation, which can lead to repeated exploration, inconsistent decisions, or failure to recover task-relevant information. These challenges suggest that long-horizon embodied agents require not only longer context, but persistent, structured, and dynamically updated memory.

Existing memory-augmented embodied agents have explored Retrieval-Augmented Generation (RAG) to improve long-term planning \citep{tan2024cradle,glocker2025llm,wang2023voyager,agashe2024agent,fu2024msi,zhao2024expel,chen2024automanual}. However, most of these systems are designed for simulated environments or rely on simple textual memory such as a single RAG or buffer, which is insufficient for capturing the spatial structure and dynamic complexity of the physical world. Moreover, many multi-module memory systems update their components sequentially, introducing additional inference latency and limiting their applicability to real-time robotic interaction.

To address these challenges, we propose \textbf{RoboMemory}, a hierarchical memory framework for long-horizon embodied agents. Inspired by the layered organization of human memory, RoboMemory decomposes memory into sensory, short-term, and long-term components, enabling the agent to process transient observations, maintain task-relevant context, and accumulate persistent experience over time. More importantly, RoboMemory introduces a comprehensive embodied memory module that jointly maintains temporal, spatial, semantic, and episodic information. These memory components are updated in parallel, allowing the agent to preserve a coherent and spatially grounded representation of the environment while reducing memory-update latency.

RoboMemory follows a closed-loop embodied intelligence paradigm that couples memory construction, memory retrieval, planning, and real-world execution. Instead of relying solely on the current observation, the agent continuously updates multiple forms of memory from its interactions and reuses accumulated experience to guide future decisions. This design allows RoboMemory to adapt its behavior in dynamic environments and improve performance across tasks through interactive learning.

We evaluate RoboMemory on EmbodiedBench, a challenging benchmark for long-horizon embodied planning \citep{yang2025embodiedbench}. Using the open-source Qwen2.5-VL-72B-Ins \citep{bai2025qwen2} as the base model, the full
RoboMemory system improves the average success rate by 26.5 percentage points over the single-agent Qwen2.5-VL-72B-Ins baseline. Notably, despite relying on an open-source VLM backbone, RoboMemory achieves performance comparable to strong closed-source VLM agents such as Claude-3.5-Sonnet. In real-world deployments, RoboMemory further demonstrates memory reuse across repeated interactions, leveraging accumulated experience from previous attempts to guide subsequent task execution.

Our main contributions are threefold:
\begin{itemize}
    \item We propose RoboMemory, an online multi-memory architecture for embodied agents, where temporal, spatial, semantic, and episodic memories are updated and retrieved in parallel to maintain task-relevant context during long-horizon interaction.

    \item We design a retrieval-based \textbf{incremental update algorithm} for the real-time evolution of Spatial Knowledge Graphs (KGs). This algorithm bridges the gap between static knowledge representation and dynamic embodied interaction, establishing KGs as a practical and scalable modality for Spatial Memory. By retrieving relevant subgraphs, detecting local inconsistencies, and merging new observations, our method ensures efficient maintenance and overcomes the scalability bottlenecks that previously hindered KG-based approaches.
    
    \item RoboMemory supports \textbf{interactive environmental learning in real-world environments}. It enables continued interaction without memory reset, allowing accumulated experience to improve future task execution, with real-world experiments demonstrating the reuse of accumulated memory across repeated task attempts.
\end{itemize}

\section{RoboMemory}

\begin{figure*}[ht]
    \centering
    \vspace{-5pt}
    \includegraphics[width=0.99\linewidth]{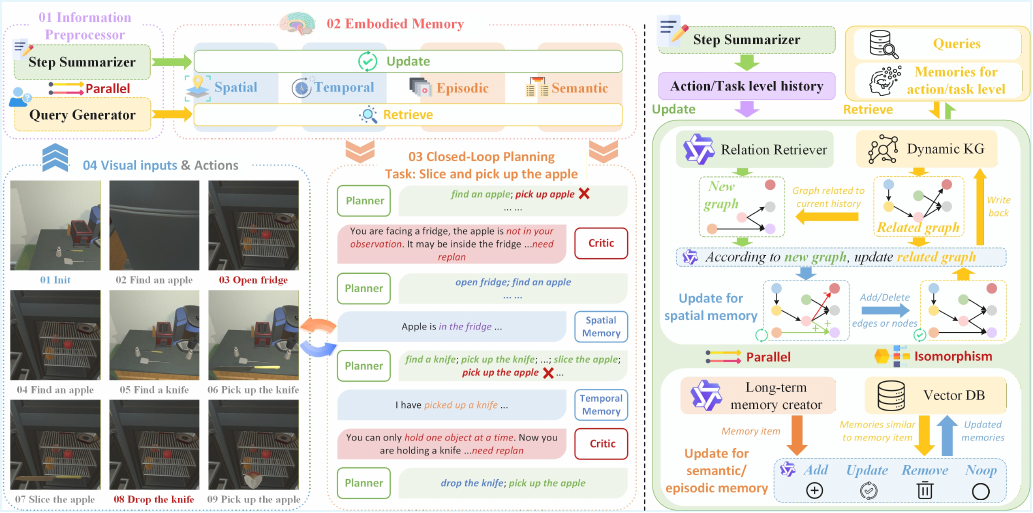}
    \caption{
        (a) Left: The loop where the Planner, Critic, and Comprehensive Embodied Memory interact to adjust plans based on real-time visual inputs.  (b) Right: The update mechanism of Spatial memory, Semantic memory, and Episodic memory.
    }
    \label{fig:main}
    \vspace{-16pt}
\end{figure*}

RoboMemory is a hierarchical embodied agent system that equips robots with three core memory capabilities: historical interaction logs, dynamically updated spatial layouts, and accumulated task knowledge. As illustrated in Figure~\ref{fig:main}, in each iteration, RoboMemory follows a ``Perception -- Memory Update -- Memory Retrieval -- Planning -- Execution" pipeline, ensuring that the agent continuously calibrates its memory and behavior in dynamic environments.

First, the information preprocessor converts multimodal sensor inputs into a textual summary of the current scene, which serves as the primary input to the Comprehensive Embodied Memory. Next, the Comprehensive Embodied Memory updates its internal representations, including action histories, object locations, and experiential knowledge. After the memory is updated, the memory system retrieves contextually relevant entries to inform the Closed-Loop Planning Module. Then, leveraging this contextual memory, the Closed-Loop Planning Module generates high-level, text-based action instructions. Finally, these commands are dispatched to low-level executors, which directly control the robot and complete the instructions. The execution process is demonstrated in Algorithm~\ref{alg:robomemory} in the appendix.

\subsection{Information Preprocessor}
\label{sec:information_preprocessor}

At each time step \(t\), RoboMemory receives a visual observation \(\mathcal{O}_t\): an RGB frame (in simulation) or a short video clip (on physical robots), representing the agent’s observations. Since raw visual data is unsuitable for direct use in memory construction and retrieval, RoboMemory first employs an information preprocessor to convert multimodal observations into textual representations, thereby providing a semantic interface for subsequent memory and planning modules. The information preprocessor executes two Vision-Language Models (VLMs) in parallel: (1) \textit{Step summarizer} \(\mathcal{S}\): It transforms \(\mathcal{O}_t\) into a concise textual description \(s_t\) of the just-executed action. The string \(s_t\) is stored in the system’s working memory. (2) \textit{Query generator} \(\mathcal{Q}\): It derives a list of queries \(q_t = [q_t^{(1)}, q_t^{(2)}, \dots, q_t^{(N)}]\) from the same observation \(\mathcal{O}_t\). Each query \(q^{(i)}_t\) is a natural-language question targeting information potentially useful for planning.

Together, \(\mathcal{S}\) and \(\mathcal{Q}\) provide a lightweight text-based interface between raw sensory inputs and RoboMemory's Comprehensive Embodied Memory System.

\subsection{Comprehensive Embodied Memory}

To address the long-term memory limitations in current embodied agent frameworks, we propose the Comprehensive Embodied Memory System. This system consists of multiple memory modules. We denote the memory system containing \(L\) distinct modules as \(M_t = [M_t^{(1)}, M_t^{(2)}, ..., M_t^{(L)}]\), where \(M_t\) represents the memory stored at step \(t\), and \(M_t^{(l)}\) denotes the \(l\)-th memory module. Generally, the update and retrieval process at iteration \(t\) of a general memory is shown below:

\vspace{-3pt}

\begin{equation}
M_t = \mathcal{U}(M_{t -1}, s_t)
\end{equation}

\vspace{-10pt}

\begin{equation}
r_t = \mathcal{R}(M_t, q_t)
\end{equation}

\vspace{-3pt}

First, we update memory modules with the memory update algorithm \(\mathcal{U}\), where we update the previous memory \(M_{t-1}\) using the latest summarization \(s_t\). Then, with updated \(M_t\), we use the memory retrieve algorithm \(\mathcal{R}\) to retrieve the information that is useful for the planning module. In \(\mathcal{R}\), we use queries \(q_t\) to query \(M_t\), yielding retrieval results from each module: \(r_t = [r_t^{(1)}, r_t^{(2)}, ..., r_t^{(L)}]\). These results are then passed to the planning module, which helps it plan future movements. However, sequentially updating and retrieving from \(L\) modules incurs high latency. Therefore, we parallelize these steps across all modules, which significantly enhances the efficiency of the memory system.

Functionally, inspired by cognitive psychology \cite{liu2025advances}, RoboMemory organizes memory into modules with different roles. The step summarizer first condenses the agent's interaction with the environment at each iteration, capturing the most immediate observations and actions. Based on these summaries, Temporal Memory records the agent's action history over time, while Spatial Memory dynamically maintains the spatial relationships among objects in the environment. Together, they provide the Closed-Loop Planner with a detailed and continuously updated task context for future planning. Beyond the current task, Semantic Memory and Episodic Memory are updated when meaningful information emerges, such as after task completion. They store abstract factual, event-based, and experiential knowledge distilled from past interactions, enabling RoboMemory to reuse prior experience and improve future task performance. We now detail each module.

\subsubsection{Spatial and Temporal Memory}

\textbf{Temporal Memory.} In Temporal memory, we record interactions between the robot and the environment (i.e., Sensory Memory) of each iteration sequentially. This memory maintains a sequential record of recent action history.  In a specific design, temporal memory is a sequential buffer with a capacity of  \(N\) interaction summaries, each generated by an information preprocessor. When the buffer is full, we compress \(N\) steps into a single summarized entry using a VLM, which is then reinserted at the front of the buffer, ensuring continuous context retention without unbounded growth. This recursive compression trades fine-grained details for bounded memory size. During retrieval, the current temporal buffer is serialized as textual context for downstream modules.

\textbf{Spatial Memory.} The spatial memory is designed to dynamically record the high-level spatial relationships of different entities in the environment. We use a dynamic KG to store high-level spatial information: objects and positions in the environment serve as KG vertices, and spatial relations between them are encoded as edges. The KG focuses on high-level spatial relations (e.g., ``cup on table", ``key left of drawer"). With this design, the spatial KG focuses on semantically meaningful, task-relevant relations. This spatial information enhances the agent’s spatial reasoning capability in dynamic environments. 

However, as related work shows, most KG construction algorithms are designed for offline scene representations. This does not meet the demand of using KG as spatial memory for the agent. Instead, the KG must be updated efficiently in response to new observations. To address this issue, we introduce a retrieval-driven, incremental KG update algorithm that maintains a locally modifiable, globally consistent, and dynamically adaptive spatial memory. As illustrated in the right panel of Figure~\ref{fig:main}, the update process proceeds in four steps: (1) retrieves the most relevant sub-KG around new observations. (2) Injects new relations from the current observation by a VLM-based Relation Retriever. (3) Detects and resolves conflicts between newly extracted relations and existing ones (e.g., ``cup on table" vs. ``cup in drawer") using a VLM-based resolver, which decides whether to add, delete, or modify edges. (4) Merges back and prunes isolated vertices. Moreover, our retrieval-based incremental update algorithm is accompanied by provable efficiency guarantees. For a KG with \( n \) vertices and maximum degree \( D \), the number of vertices processed per update is bounded by \( O(D^K) \), where \( K \) is the retrieval hop distance (see Appendix~\ref{sec:proof} for analysis).  Further implementation details are provided in Algorithm~\ref{alg:KG-update} in the appendix. 

\subsubsection{Semantic and Episodic Memory}

Semantic and Episodic memory serve as long-term memory modules that retain reusable information beyond the current interaction. 

Following cognitive psychology, Semantic memory stores relatively time-independent facts, such as task-relevant environmental knowledge and reusable execution constraints, whereas Episodic memory stores autobiographical records of the agent's past task attempts, capturing what the agent did, what it observed, and what outcomes it received during previous interactions.

RoboMemory implements both modules with a shared vector-database-based memory management mechanism. Each memory item is represented as a natural-language description and embedded into a semantic vector for similarity-based retrieval. 

The database is updated only when new long-term information is detected, such as after completing a subtask or encountering important environmental feedback. During an update, a Long-term Memory Creator first generates candidate memory items from the current short-term context. For each candidate item, the system retrieves the top-\(S\) most similar existing items via cosine similarity. A VLM-based updater then compares the candidate with the retrieved items and decides whether to \textit{add} the new item, \textit{update} an existing item, \textit{remove} outdated information, or perform \textit{no-op} when the candidate is redundant. 
Since each update only involves at most \(S\) retrieved items, the update cost remains bounded with respect to the database size.

The two modules share the same storage and update mechanism but retain different types of long-term information: Semantic memory abstracts stable knowledge from interactions, while Episodic memory preserves concrete task-level experiences. 
At retrieval time, both modules return the top-\(N\) relevant memory items to downstream planning modules.

\subsection{Closed-Loop Planning Module and Low-Level Executor}

\textbf{The Closed-Loop Planning Module} integrates information provided by Temporal, Spatial, Semantic and Episodic memory, and current observations to perform action planning. Each action is planned and passed on to the low-level executor for execution. To enable closed-loop control in embodied environments, the Closed-Loop Planning Module adopts the Planner-Critic mechanism \citep{lei2025clea}, which consists of the planner and the critic module powered by the same VLM model with different prompts. We denote the planner module as \(\mathcal{P}\), while the critic module as \(\mathcal{C}\).  The demonstration of this process is shown in Figure~\ref{fig:main}.

\textbf{The Low-Level Executor} is responsible for grounding the planned actions into executable low-level operations. Together with the high-level planner, it forms a two-layer hierarchical agent framework, enabling RoboMemory to accomplish longer-horizon tasks in real-world environments. We employ a LoRA-finetuned VLA model, \(\pi_0\) \citep{hu2021lora, black2024pi_0}, to generate manipulation actions, and a SLAM-based navigation model for locomotion. 


\begin{table*}[ht]
\caption{Comparison of Success Rates (SR) and Goal Condition Success Rates (GC) across difficulty levels (Base/Long) on EB-ALFRED and EB-Habitat benchmarks. Values are reported in percentages (\%).}
\label{tab:combined_full_benchmarks}

\vspace{-5pt}

\centering
\scriptsize
\setlength{\tabcolsep}{5pt} 
\begin{tabular}{l|l|cc|cccc|cccc}
\toprule
\multirow{3}{*}{Method} & \multirow{3}{*}{Type} &
\multicolumn{2}{c}{} & 
\multicolumn{4}{c}{EB-ALFRED} & 
\multicolumn{4}{c}{EB-Habitat} \\
\cmidrule(lr){5-8} \cmidrule(lr){9-12} 

& & 
\multicolumn{2}{c}{Average} & 
\multicolumn{2}{c}{Base} & \multicolumn{2}{c}{Long} & 
\multicolumn{2}{c}{Base} & \multicolumn{2}{c}{Long} \\
\cmidrule(lr){3-4} \cmidrule(lr){5-6} \cmidrule(lr){7-8} \cmidrule(lr){9-10} \cmidrule(lr){11-12}

& & SR & GC & SR & GC & SR & GC & SR & GC & SR & GC \\
\midrule

\multicolumn{12}{l}{\textit{Single VLM-Agents}} \\
\midrule
GPT-4o & \multirow{7}{*}{Closed-source} & 67.0 & \underline{74.9} & 64.0 & 74.0 & 54.0 & \underline{62.5} & 86.0 & 90.7 & \textbf{64.0} & \underline{72.2} \\
GPT-4o-mini & & 30.5 & 40.9 & 34.0 & 47.8 & 0.0 & 17.0 & 74.0 & 77.5 & 14.0 & 21.3 \\
Claude-3.7-Sonnet & & 68.5 & - & 68.0 & - & \textbf{70.0} & - & 90.0 & - & 46.0 & - \\
Claude-3.5-Sonnet & & \underline{69.5} & 71.8 & \textbf{72.0} & 72.0 & 52.0 & 54.5 & \textbf{96.0} & \textbf{97.5} & 58.0 & 63.3 \\
Gemini-1.5-Pro & & 68.0 & 73.3 & \underline{70.0} & \underline{74.3} & 58.0 & 65.0 & 92.0 & 92.5 & 52.0 & 61.2 \\
Gemini-2.0-flash & & 57.0 & 61.5 & 62.0 & 65.7 & 58.0 & 62.0 & 82.0 & 82.0 & 26.0 & 36.2 \\
\midrule
Llama-3.2-90B-Vision-Ins & \multirow{5}{*}{Open-source} & 40.5 & 46.6 & 38.0 & 43.7 & 16.0 & 24.0 & \underline{94.0} & \underline{94.5} & 14.0 & 24.3 \\
InternVL2.5-78B & & 47.0 & 52.9 & 38.0 & 42.3 & 42.0 & 49.0 & 80.0 & 82.0 & 28.0 & 38.2 \\
InternVL2.5-38B & & 37.5 & 42.6 & 36.0 & 37.3 & 26.0 & 36.5 & 60.0 & 61.5 & 28.0 & 35.0 \\
InternVL3-78B & & 49.5 & - & 38.0 & - & 36.0 & - & 84.0 & - & 40.0 & - \\
Qwen2.5-VL-72B-Ins & & 44.0 & - & 50.0 & - & 34.0 & - & 74.0 & - & 18.0 & - \\
\midrule
\multicolumn{12}{l}{\textit{VLM-Agent Frameworks}} \\
\midrule
Voyager (Qwen2.5-VL-72B-Ins) & \multirow{5}{*}{Baselines} & 46.5 & 66.4 & 56.0 & 73.2 & 32.0 & 54.2 & 76.0 & 87.0 & 22.0 & 51.0 \\
Reflexion (Qwen2.5-VL-72B-Ins) & & 38.3 & 51.1 & 48.0 & 54.0 & 10.0 & 33.0 & 80.0 & 84.2 & 15.0 & 33.0 \\
Cradle (Qwen2.5-VL-72B-Ins) & & 44.5 & 57.0 & 54.0 & 67.9 & 32.0 & 41.0 & 62.0 & 67.0 & 30.0 & 52.1 \\
RoboOS (Qwen2.5-VL-72B-Ins) & & 25.5 & 33.0 & 32.0 & 38.4 & 12.0 & 17.6 & 38.0 & 47.8 & 20.0 & 28.2 \\
RoboOS (RoboBrain2-32B) & & 20.0 & 25.7 & 32.0 & 37.2 & 8.0 & 13.2 & 28.0 & 34.8 & 12.0 & 17.4 \\
\midrule
\textbf{RoboMemory (Qwen2.5-VL-72B-Ins)} & \textbf{Ours} & \textbf{70.5} & \textbf{79.7} & 68.0 & \textbf{75.5} & \underline{66.0} & \textbf{81.3} & 86.0 & 88.0 & \underline{62.0} & \textbf{74.0} \\
\bottomrule
\end{tabular}
\vspace{-10pt}
\end{table*}

\section{Experiments}

\subsection{Benchmarks}

To evaluate RoboMemory's task planning ability, we select a subset of the EmbodiedBench EB-ALFRED and EB-Habitat benchmark \citep{yang2025embodiedbench}. We select the Base and Long subsets because they aim to test the agent's planning ability. The Base and Long subsets of the two benchmarks comprise 200 tasks for complex embodied tasks. The EB-ALFRED and EB-Habitat benchmarks provide a visually grounded operational setting that closely mimics real-world conditions (see Appendix~\ref{sec:add_env_set} for environment details), enabling direct comparison with established baselines. Moreover, we set up a real-world environment to evaluate RoboMemory's ability to reuse accumulated experience through long-term memory during interactive environmental learning. Details about the benchmark settings are shown in Appendix~\ref{sec:add_env_set}.

\begin{figure*}[ht]
\vspace{-5pt}
\centering

\begin{minipage}{0.47\textwidth}
    \centering
    \includegraphics[width=1.0\linewidth]{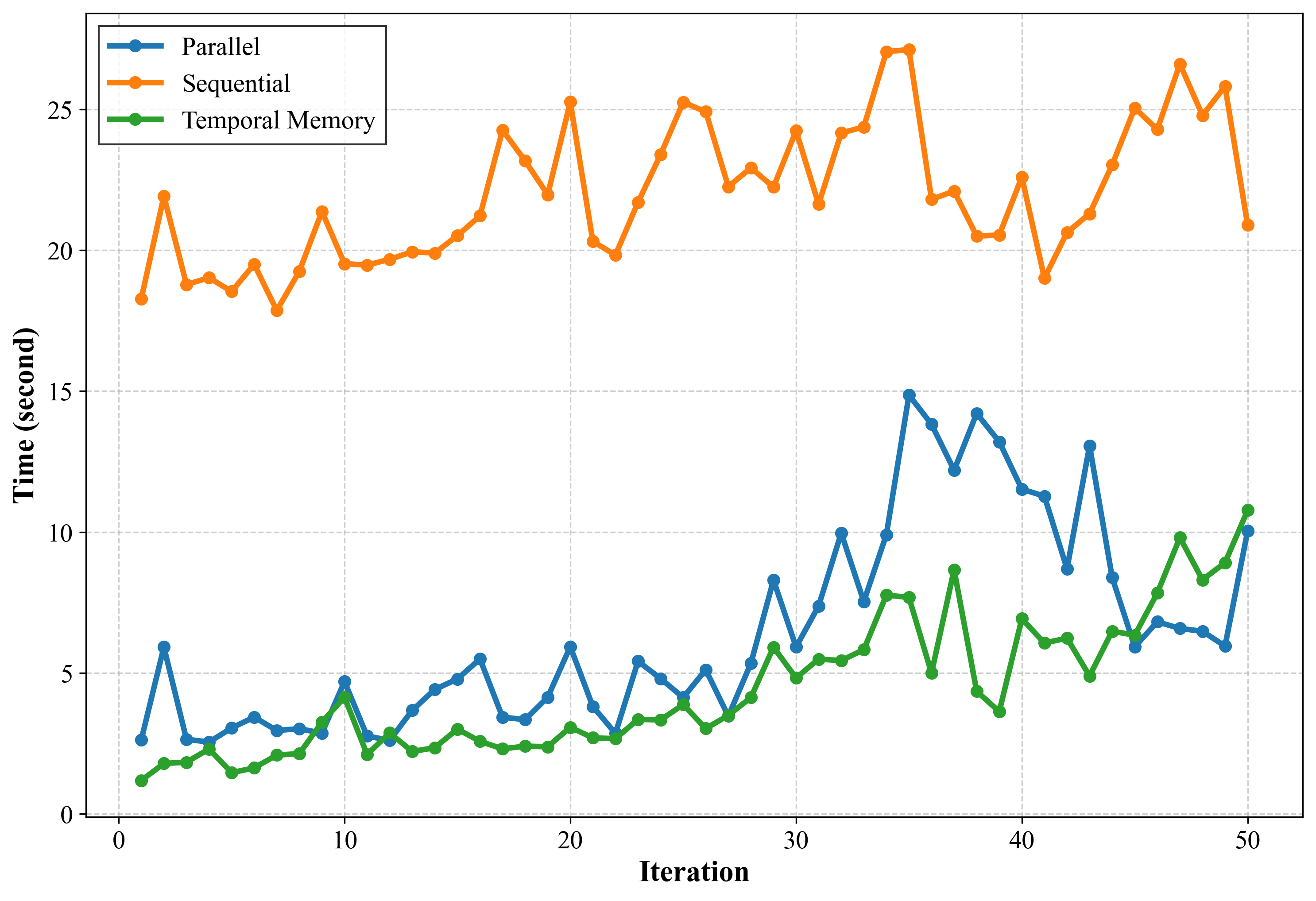}
    \vspace{-20pt}
    \caption{Efficiency improvement of Comprehensive Embodied Memory System.}
    \label{fig:eff}
\end{minipage}
\hfill
\begin{minipage}{0.5\textwidth}
    \centering
    \footnotesize
    \captionof{table}{Ablation Study on RoboMemory's Success Rate (SR).}
    \vspace{-8pt}
    \label{tab:ablation}

    \begin{tabular}{lccc}
        \toprule
        Method & Avg. & Base & Long \\
        \midrule
        \textbf{RoboMemory}
            & \textbf{67\%} & \textbf{68\%} & \textbf{66\%} \\
        w/o Critic
            & 55\% & 60\% & 50\% \\
        w/o Spatial Memory
            & 47\% & 52\% & 42\% \\
        w/o  Long-term Memory
            & 57\% & 66\% & 48\% \\
        w/o Episodic Memory
            & 62\% & 68\% & 56\% \\
        w/o Semantic Memory
            & 58\% & 66\% & 50\% \\
        \bottomrule
    \end{tabular}

\end{minipage}

\vspace{-10pt}
\end{figure*}

\subsection{Main Results}

As shown in Table~\ref{tab:combined_full_benchmarks}, RoboMemory achieves substantial improvements on both EB-ALFRED and EB-Habitat. Compared with the single-agent Qwen2.5-VL-72B-Ins baseline, the full RoboMemory system improves the average SR from 44.0\% to 70.5\%, yielding a gain of 26.5 percentage points. This demonstrates that the complete RoboMemory framework substantially enhances the long-horizon planning ability of an open-source VLM backbone.

Notably, RoboMemory also closes the gap between open-source and closed-source VLM agents. RoboMemory achieves an average SR comparable to Claude-3.5-Sonnet (70.5\% vs. 69.5\%), while achieving a 7.9 percentage-point higher average GC. Furthermore, RoboMemory outperforms existing VLM-agent frameworks using the same backbone.

We attribute these gains to two key factors: the comprehensive memory system provides persistent and structured contextual information, while the Planner-Critic mechanism enables closed-loop replanning when failures or unexpected situations occur. These results suggest that, with structured memory and closed-loop planning, an agent framework built on an open-source VLM can achieve performance comparable to strong closed-source VLM agents.

RoboMemory is also more robust to longer task horizons. On EB-Habitat, the success rate drops by 24 percentage points from Base to Long, compared with 56 points for Qwen2.5-VL-72B-Ins and 38 points for Claude-3.5-Sonnet. On EB-ALFRED, RoboMemory drops by only 2 points.

\subsection{Efficiency Analysis}

To evaluate the efficiency of the Comprehensive Embodied Memory module, we tested RoboMemory by executing 10 long-horizon tasks. All testing was conducted under the same hardware and software settings. We measure the aggregate wall-clock time consumed by memory update and retrieval operations per high-level decision. We analyzed the scaling behavior of memory update latency across three distinct configurations: (1) fully parallel update and retrieval across all memory modules; (2) sequential update of each memory module without parallelism; and (3) update of only the most fundamental memory component: Temporal memory. Results are presented in Figure~\ref{fig:eff}. As shown, our parallel update strategy enables updating a multi-module memory system with latency comparable to that of updating a single base memory module. This demonstrates the critical efficiency gains afforded by parallelization across the memory architecture.

\begin{figure*}[ht]
\vspace{-8pt}
  \centering
  \begin{minipage}[t]{0.45\textwidth}
    \centering
    \includegraphics[width=0.95\linewidth]{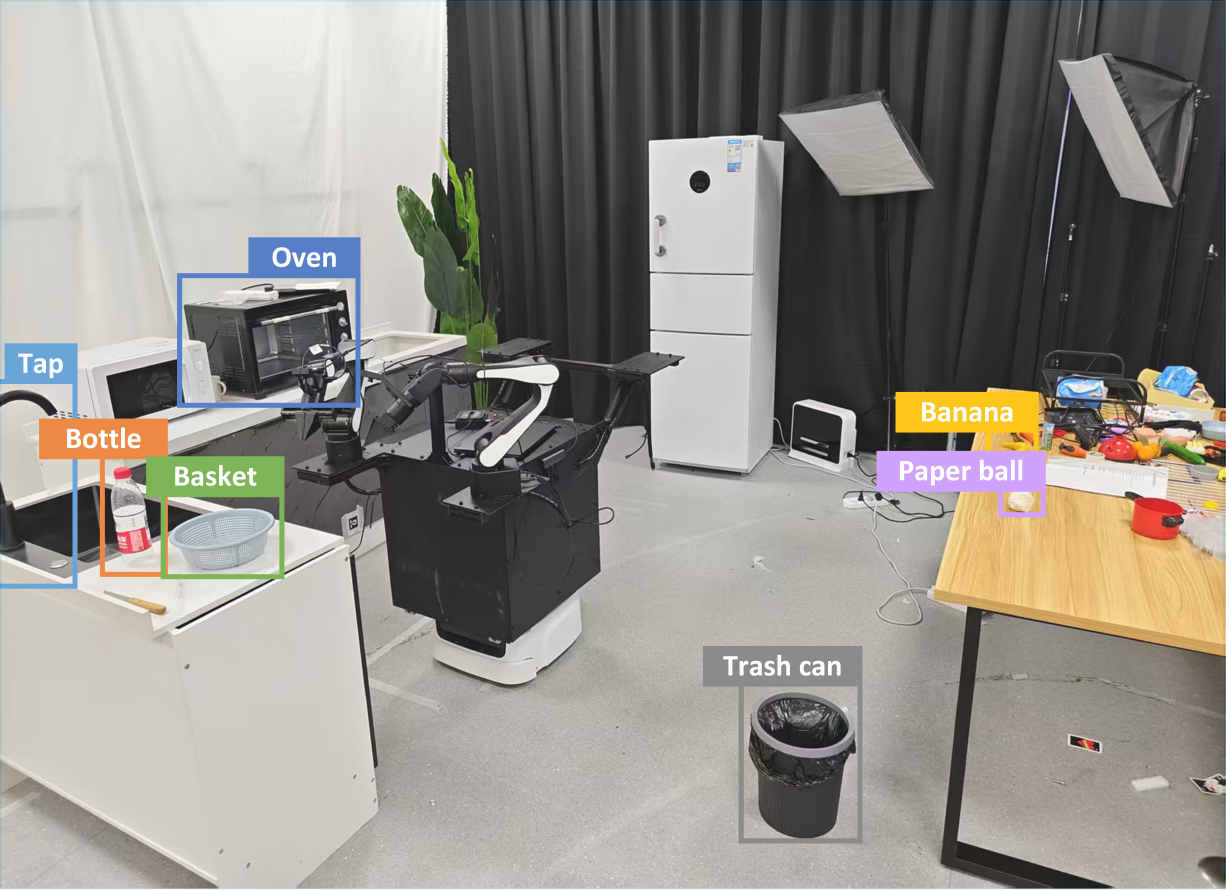}
    \caption{Visualization of the experimental environment.}
    \label{fig:env}
  \end{minipage}
  \hfill
  \begin{minipage}[t]{0.5\textwidth}
    \centering
    \includegraphics[width=0.99\linewidth]{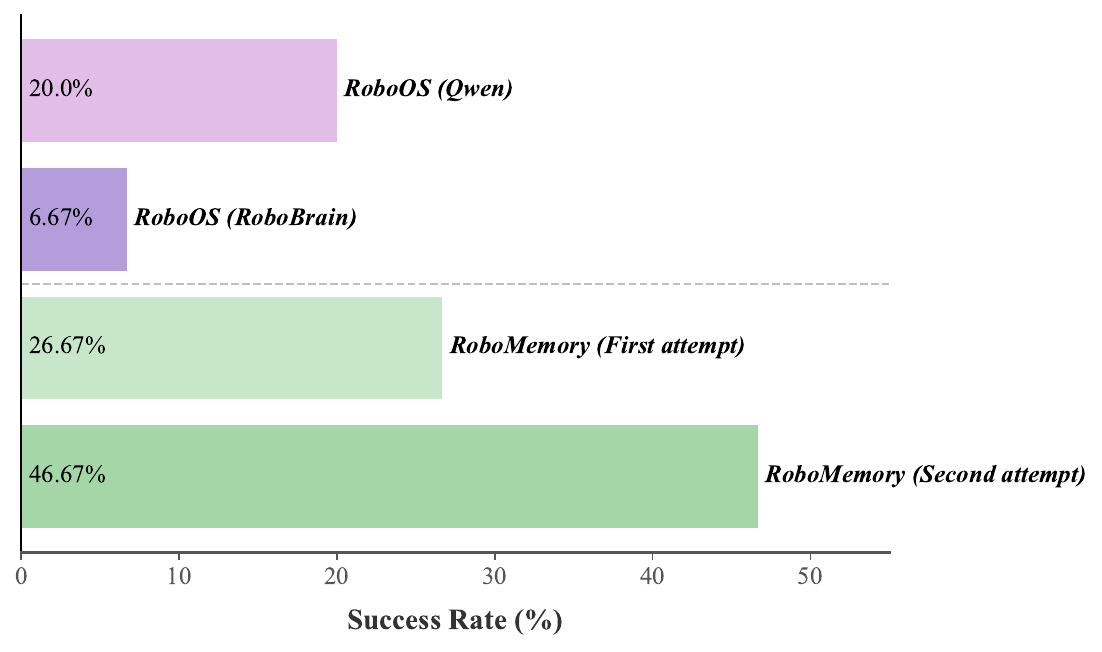}
    \caption{Real-world experiment results. ``Qwen" denotes Qwen2.5-VL-72B-Ins; ``RoboBrain" denotes RoboBrain2.0-32B.}
    \label{fig:real_result}
  \end{minipage}
  \vspace{-20pt}
\end{figure*}

\subsection{Ablation Studies}

We used the full Base and Long Subset from EB-ALFRED to validate RoboMemory's effectiveness.
We removed each component systematically and observed performance changes across task categories. We use the success rate as our metric. Results are shown in Table~\ref{tab:ablation}.

\textbf{Long-term Memory:} Adding long-term memory significantly improves RoboMemory's success rate. The experiment shows that it enables interactive environmental learning while attempting to complete tasks. Semantic memory learns the properties of low-level skills, such as in what circumstances an action may fail, while Episodic memory records all task attempts, providing valuable experience at the task level and giving insight into how to complete a task successfully. This helps RoboMemory predict the outcomes of actions and avoid ineffective attempts. This ability indicates that RoboMemory has an interactive environmental learning capability. We further ablate Semantic memory and Episodic memory separately. The results show that both modules contribute to RoboMemory's overall performance.

\textbf{Spatial Memory:} Spatial memory is crucial for embodied agents, especially given that current pretrained VLMs have limited spatial understanding ability. Our novel dynamic KG update algorithm enables KG-based spatial memory in dynamic environments. This spatial reasoning helps RoboMemory handle partially observable embodied settings.

\textbf{Critic Module:} Table~\ref{tab:ablation} shows performance without the critic module (55\% vs 67\% with full system). This drop highlights how the critic's closed-loop planning adapts to dynamic environments. It helps RoboMemory recover from failures faster and handle unexpected situations better.

Additional matched controls disentangling the contributions of the Planner--Critic mechanism and the dynamic Spatial KG are provided in Appendix~\ref{sec:matched_controls}.

\subsection{Real-World Robot Deployment}

To study how RoboMemory supports interactive environmental learning in the real world, with a focus on accumulated experience reuse, we designed a kitchen environment inspired by EB-ALFRED and EB-Habitat. The environment is shown in Figure~\ref{fig:env}. Additional experiment details are in Appendix \ref{sec:add_env_set}.

To evaluate how accumulated experience supports interactive environmental learning, we ran each task twice without clearing long-term memory between attempts. Meanwhile, we compare RoboMemory with RoboOS, a strong prior embodied-agent framework, as the real-world baseline. The success rates for the first and second attempts and different settings of RoboOS are shown in Figure~\ref{fig:real_result}.

The second attempt showed noticeably higher success rates. This suggests that accumulated experience retained in long-term memory can support subsequent task attempts in real embodied environments. Key observations include: (1) Closed-loop error recovery: RoboMemory retries failed actions when possible, even if the low-level executor (VLA model) fails. (2) Spatial reasoning: RoboMemory remembers object locations and spatial relationships using its memory. (3) Interactive environmental learning: RoboMemory analyzes failure causes reasonably. These analyses guide future decisions. Detailed examples demonstrating these capabilities and further discussions are provided in Appendix~\ref{sec:case}.

Moreover, we observe a significant drop in task success rates when deploying the agent with the Low-level Executor in real-world environments. This performance degradation primarily stems from the executor's inherent limitations: (1) The VLA model exhibits unreliable instruction-following capabilities, frequently failing during grasping actions or selecting incorrect objects; (2) Pre-trained VLM models demonstrate inadequate video understanding capability. They struggle to interpret dynamic visual information such as action failures or state changes. These limitations collectively contribute to the reduced performance compared to simulated environments.

\section{Conclusion}

This paper proposes RoboMemory, a brain-inspired multi-memory framework that facilitates long-horizon planning and interactive environmental learning in real-world embodied systems. Experiments on EmbodiedBench demonstrate that RoboMemory improves an open-source VLM backbone and achieves performance comparable to strong closed-source VLM agents, while ablation studies confirm the critical roles of the Critic module and spatial/long-term memory. Real-world deployment further supports interactive learning by demonstrating that accumulated experience retained in long-term memory can be reused across repeated task attempts. Despite limitations arising from reasoning errors and executor dependence, RoboMemory provides a foundation for generalizable, memory-augmented agents, with future work aimed at refining reasoning and enhancing execution robustness.

\section{Limitations}

This work explores how to construct a memory system for embodied agents. However, RoboMemory still relies on two limiting assumptions: pretrained VLMs can reliably interpret observations, and low-level executors can faithfully ground high-level language instructions. In practice, failures may arise from incorrect perception, hallucinated state changes, imperfect video understanding, or invalid plans, especially when execution details such as precise grasp poses are difficult to specify textually. Our real-world evaluation is also limited to a kitchen-like setting with repeated task attempts. Future work may improve robustness by developing richer multimodal planner-executor interfaces, stronger failure detection, and broader evaluations across diverse robots and environments. In addition, the current Spatial Memory represents high-level relational structure rather than metric 3D geometry, leaving multi-view geometric grounding for future work.

\acknowledgments{
This work was supported by NSFC with Grant No. 62293482, by the Basic Research Project No. HZQB-KCZYZ-2021067 of Hetao Shenzhen-HK S\&T Cooperation Zone, by NSFC with Grant No. 62573371 and Grant No. 62501514, by the Shenzhen General Program No. JCYJ20220530143600001, by the Shenzhen Municipal-Level Talent Program, by the Shenzhen Outstanding Talents Training Fund 202002, by the Guangdong Research Project No.2017ZT07X152 and No. 2019CX01X104, by the Guangdong Provincial Key Laboratory of Future Networks of Intelligence (Grant No. 2022B1212010001), by the Guangdong Provincial Key Laboratory of BigData Computing CUHK-Shenzhen, by the NSFC 61931024\&12326610, by the Key Area R\&D Program of Guangdong Province with grant No. 2018B030338001, by the Shenzhen Key Laboratory of Big Data and Artificial Intelligence (Grant No. SYSPG20241211173853027), by National Key Research and Development Program of China: 2025YFF0515300 and 2025YFF0515304, by China Association for Science and Technology Youth Care Program, by Key Laboratory of Tibetan Information Processing, Ministry of Education, Xining, 810008, by Qinghai Provincial Key Laboratory of Tibetan Information Processing and Machine Translation, Xining, 810008, Tibetan Information Processing Engineering Technology Research Center of Qinghai Province, Xining, 810008, by the Shenzhen-Hong Kong Joint Funding No. SGDX20211123112401002, by Tencent \& Huawei Open Fund, and by Guangdong Province Radio Science Data Center under Grant 2025B1212070001.
}

\bibliography{references}

\clearpage
\newpage

\appendix

\section{Related Work}
\label{sec:add_rel}

\begin{table*}[htbp]
\centering
\scriptsize
\setlength{\tabcolsep}{3pt} 
\caption{Comparison of Memory-Related Methods in Embodied Agent}
\label{tab:memory_comparison}
\begin{tabular}{lcccccclc}
\toprule
\textbf{Method} & \textbf{Multimodal} & \textbf{Episodic} & \textbf{Semantic} & \textbf{Spatial} & \textbf{Temporal} & \textbf{Procedural} & \textbf{Memory Implementation} & \textbf{Real Robot} \\
\midrule

NeSyC \citep{choi2025nesyc} & \checkmark & & \checkmark & & \checkmark & & Symbolic logic rules & \checkmark \\
Reflexion \citep{shinn2024reflexion} & &  & \checkmark & & \checkmark & & Buffer & \\
Voyager \citep{wang2023voyager}  & & & & & & \checkmark & RAG & \\
MSI-Agent \citep{fu2024msi} & & & \checkmark & & \checkmark & & Database, RAG & \\
CoELA \citep{zhang2023building} & \checkmark & \checkmark & \checkmark & & & \checkmark & Top-down semantic map &  \\
Cradle \citep{tan2024cradle} & \checkmark & \checkmark  & & & \checkmark & \checkmark & RAG & \\
Agent-S \citep{agashe2024agent} & \checkmark & \checkmark & \checkmark & & \checkmark & & RAG & \\
Expel \citep{zhao2024expel} & & & \checkmark & & \checkmark & & Buffer & \\
AutoManual \citep{chen2024automanual} & & \checkmark & \checkmark & & & \checkmark & Buffer & \\
HiRobot \citep{shi2025hi} & \checkmark & & & & & & / & \checkmark \\
Being-0 \citep{yuan2025being} & \checkmark & \checkmark & & & \checkmark & & Buffer & \checkmark \\ 
RoboOS \citep{tan2025roboos} & \checkmark &  & & \checkmark & \checkmark & & Scene graph, database & \checkmark 
\\
\midrule
RoboMemory (Ours) & \checkmark & \checkmark & \checkmark & \checkmark & \checkmark & & RAG,KG & \checkmark \\
\bottomrule
\end{tabular}

\smallskip
\end{table*}

\subsection{VLM/LLM-based Agentic Frameworks in Embodied Tasks}

The rapid advancement of VLMs/LLMs has led to diverse agent frameworks in embodied environments \citep{yao2022react,song2023llm,lin2024swiftsage}. Embodied tasks involve partial observability and long-horizon planning, requiring memory systems to retain context. Some use time-ordered context buffers for short-term memory (due to VLMs/LLMs' limited long-context processing) \citep{yao2022react,packer2023memgpt}; others adopt experience buffers as long-term semantic memory \citep{fu2024msi,shinn2024reflexion}. For long-duration tasks, skill libraries serve as procedural memory, with agents accumulating skills via interaction \citep{wang2023voyager,tan2024cradle}. However, in real-world settings, the low-level executor may fail to complete the task, making it challenging to construct a reusable, code-based skill library. So, explicit procedural memory still needs to be improved in real-world settings. Moreover, Recent efforts integrate diverse memories \citep{zhang2023building,tan2024cradle,agashe2024agent} but focus on virtual/GUI environments, leaving real-world multi-modal memory support for long-term planning under-explored.

\subsection{Spatial Memory}

Spatial information is critical for embodied agents. Prior work has explored three main representations, each with inherent limitations that motivate our high-level abstractions: \textbf{2D Semantic Maps} encode environments as pixel grids \cite{zhang2023building}. As the semantic maps are flattened, they discard 3D relationships (e.g., ``above/beneath'') and precise geometry, limiting expressiveness for manipulation tasks requiring fine-grained spatial reasoning. \textbf{3D Point Clouds} preserve accurate 3D information \cite{zhang2025nava}. However, a 3D point cloud only encodes low-level spatial information. We need to put in extra effort to help agents understand 3D Point Clouds. \textbf{Semantic Scene Graphs} \cite{tan2025roboos, loo2025open, gu2024conceptgraphs, yang20253d} offer a \textit{compact, high-level} representation directly amenable to planning. However, existing methods suffer from two key drawbacks: (1) Most previous methods adopt rigid tree-like structures, which are easy to update with new information, but cannot model non-hierarchical or cyclic spatial configurations. (2) Some previous methods try to use general graphs to store various information \cite{gutierrez2024hipporag}.  However, their design often lacks efficient dynamic update mechanisms—adding or modifying relations typically requires expensive full-graph recomputation, creating a scalability bottleneck for real-time embodied interaction.

For clarity, we summarize the differences among different memory systems in the previous work. The comparison is shown in Table~\ref{tab:memory_comparison}.

\subsection{Memory Systems in Embodied Intelligence: From Cognitive Theories to Robotic Implementations}

Embodied agents depend on memory to integrate experiences, resolve partial observability, and support long-horizon planning—capabilities rooted in cognitive psychology. This section links foundational memory theories to robotic implementations, highlighting gaps addressed by subsequent work.

\textbf{Cognitive Psychology-Inspired Memory Frameworks.}
Cognitive psychology provides hierarchical memory models for agent design. The multi-store model \cite{atkinson1968human} formalizes sensory, short-term, and long-term memory tiers. Tulving \cite{tulving1972episodic, Tulving1983-TULEOE} refined long-term memory into episodic (event-specific) and semantic (factual) subtypes. The working memory model \cite{baddeley2020working} extends short-term memory with specialized subcomponents for active information manipulation.  While classical theories describe how memory is organized, recent work on machine memory intelligence emphasizes broader capabilities such as associative representation, continual learning, and collaborative reasoning~\cite{zheng2025machine}. In embodied robotics, however, many frameworks still lack an integrated architecture across memory types and timescales. Early works \cite{choi2025nesyc, zhao2024expel} lacked full sensory-short-long-term structures, leading to inefficient knowledge accumulation. Recent frameworks \cite{agashe2024agent, tan2024cradle} incorporated episodic and semantic memory but treated them as independent buffers, ignoring dynamic tier interactions \cite{mcclelland1995there} observed in human memory. This disconnect between cognitive theories and robotic implementations limits agents' ability to adapt to real-world dynamic environments.

\textbf{Multi-Modal Memory in Embodied Agents.}
Contemporary frameworks face real-world limitations. Voyager \cite{wang2023voyager} used code-based skill libraries that fail to generalize to executor errors. Reflexion \cite{shinn2024reflexion} lacked spatial memory and relied on sequential updates, incurring high latency. RoboOS \cite{tan2025roboos} integrated scene-graph spatial memory but omitted episodic memory and suffered from rigid graph structures. Existing multi-module memory systems \cite{fu2024msi, chen2024automanual} process updates sequentially, leading to inference delays incompatible with real-time interaction. This organization is inspired by classical studies of sensory memory \cite{sperling1960information}. RoboMemory similarly processes updates across specialized memory components in parallel to reduce inference latency. Additionally, most frameworks \cite{tan2024cradle, agashe2024agent} are limited to simulated environments, lacking support for real-world multi-modal memory integration.

\begin{algorithm*}[ht]
\caption{RoboMemory Execution Process}
\label{alg:robomemory}
\begin{algorithmic}[1]
\REQUIRE Task description $\mathcal{T}$, Initial observation $\mathcal{O}_0$, Max steps $T_{max}$
\REQUIRE Modules: \text{Step Summarizer} $\mathcal{S}, \text{Query Generator} ~ \mathcal{Q}$; Memory $\mathcal{U}, \mathcal{R}$; Planner $\mathcal{P}$; Critic $\mathcal{C}$; Executor $\mathcal{E}$
\STATE \textbf{Initialize:} Global step $t \leftarrow 0$, Memory $M_t \leftarrow \emptyset$
\STATE \textbf{Initial Perception:}
\STATE $s_t, q_t \leftarrow \mathcal{SQ}(\mathcal{O}_t)$ \COMMENT{call the step summarizer and query generator in parallel}
\STATE $M_0 \leftarrow \mathcal{U}(\emptyset, s_0)$ \COMMENT{Initialize Memory with first observation}

\WHILE{$t < T_{max}$ \AND Task $\mathcal{T}$ not completed}
    \STATE \textbf{Retrieval Phase:}
    \STATE $r_t \leftarrow \mathcal{R}(M_t, q_t)$ \COMMENT{Parallel retrieval from $L$ memory modules}
    
    \STATE \textbf{Planning Phase:}
    \STATE $\mathbf{A} \leftarrow \mathcal{P}(r_t, \mathcal{O}_t, \mathcal{T})$ \COMMENT{Generate action sequence $\mathbf{A} = [a_1, a_2, \dots, a_K]$}
    
    \STATE \textbf{Execution Phase (Closed-Loop):}
    \FOR{$k = 1$ \TO $|\mathbf{A}|$}
        \STATE Let $a_k$ be the current action to execute
        \STATE $\textit{execute\_flag} \leftarrow \text{False}$
        
        \IF{$k = 1$}
            \STATE $\textit{execute\_flag} \leftarrow \text{True}$ \COMMENT{Skip Critic for the first action to avoid planner–critic deadlock before any environment transition.}
        \ELSE
            \STATE \COMMENT{Re-evaluate context for subsequent steps}
            \STATE $r_{curr} \leftarrow \mathcal{R}(M_t, q_t)$
            \IF{$\mathcal{C}(a_k, r_{curr}, \mathcal{O}_t, \mathcal{T})$ is \textbf{True}}
                \STATE $\textit{execute\_flag} \leftarrow \text{True}$
            \ELSE
                \STATE \textbf{break} \COMMENT{Critic rejects action; trigger re-planning}
            \ENDIF
        \ENDIF
        
        \IF{$\textit{execute\_flag }  \text{is} \, \textbf{True}$}
            \STATE $\mathcal{O}_{t+1} \leftarrow \mathcal{E}(a_k)$ \COMMENT{Execute action via low-level executors}
            \STATE $t \leftarrow t + 1$
            
            \STATE \textbf{Memory Update (Perception $\to$ Memory):}
            $s_t, q_t \leftarrow \mathcal{SQ}(\mathcal{O}_t)$  \COMMENT{Generate query and summary in parallel}
            \STATE $M_t \leftarrow \mathcal{U}(M_{t-1}, s_t)$ \COMMENT{Parallel update of all modules}
        \ENDIF
    \ENDFOR
\ENDWHILE
\end{algorithmic}
\end{algorithm*}

\section{Additional Experiments}
\label{sec:add_exp}

\subsection{Error Analysis}

\begin{wrapfigure}{r}{0.5\textwidth}
\centering
    \includegraphics[width=1.0\linewidth]{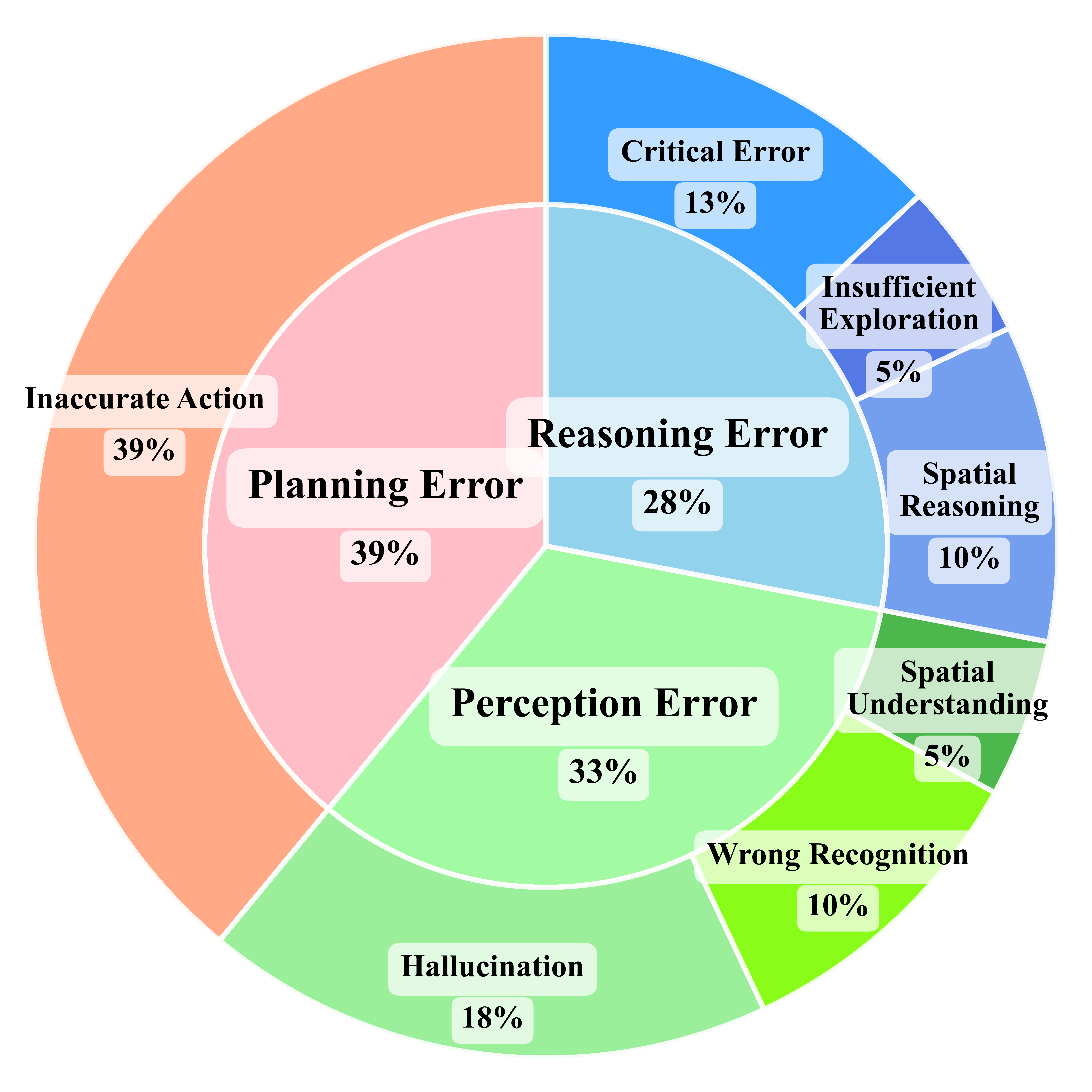}
    \caption{The reason why RoboMemory failed to complete the task.} 
    \label{fig:error_analysis}
\end{wrapfigure}

We summarize the common errors of RoboMemory in the previous experiments. We classify errors into three main types: planning errors, reasoning errors, and perception errors.

The planning errors occur when the planner fails to generate correct actions. The reasoning errors occur when the planner and critic cannot properly process input information (including current observations and memory), even when the input is correct. Perception errors occur when incorrect information is provided to the planner-critic module.

We analyze RoboMemory trajectories for failed tasks. We identify error types based on the above definitions. A single task may contain multiple errors. We calculate the occurrence probability of each error type to show RoboMemory's strengths and weaknesses. The results are shown in Figure~\ref{fig:error_analysis}. 

We can observe that among all error types, the planning errors are the most common. This means that even though the memory modules can provide comprehensive information about the RoboMemory agent's previous experience and spatial and temporal memory for the current task, the planner module may still not provide good action plans. This may be due to the capability of the pretrained base model. 

The most common perception error is the hallucination error. We can observe that although some hallucinations can be handled by the critic module or memory information, there are still some cases in which the planner ignores all insights from memory and critic and fails to complete the task.

Additionally, for EB-Habitat Long, trajectory audits further identify missed key-object recognition and repeated exploration loops as the dominant remaining failure patterns.

The detailed examples and discussions are provided in Appendix~\ref{sec:case}.

\begin{figure*}[htbp]
    \centering
    \includegraphics[width=0.99\linewidth]{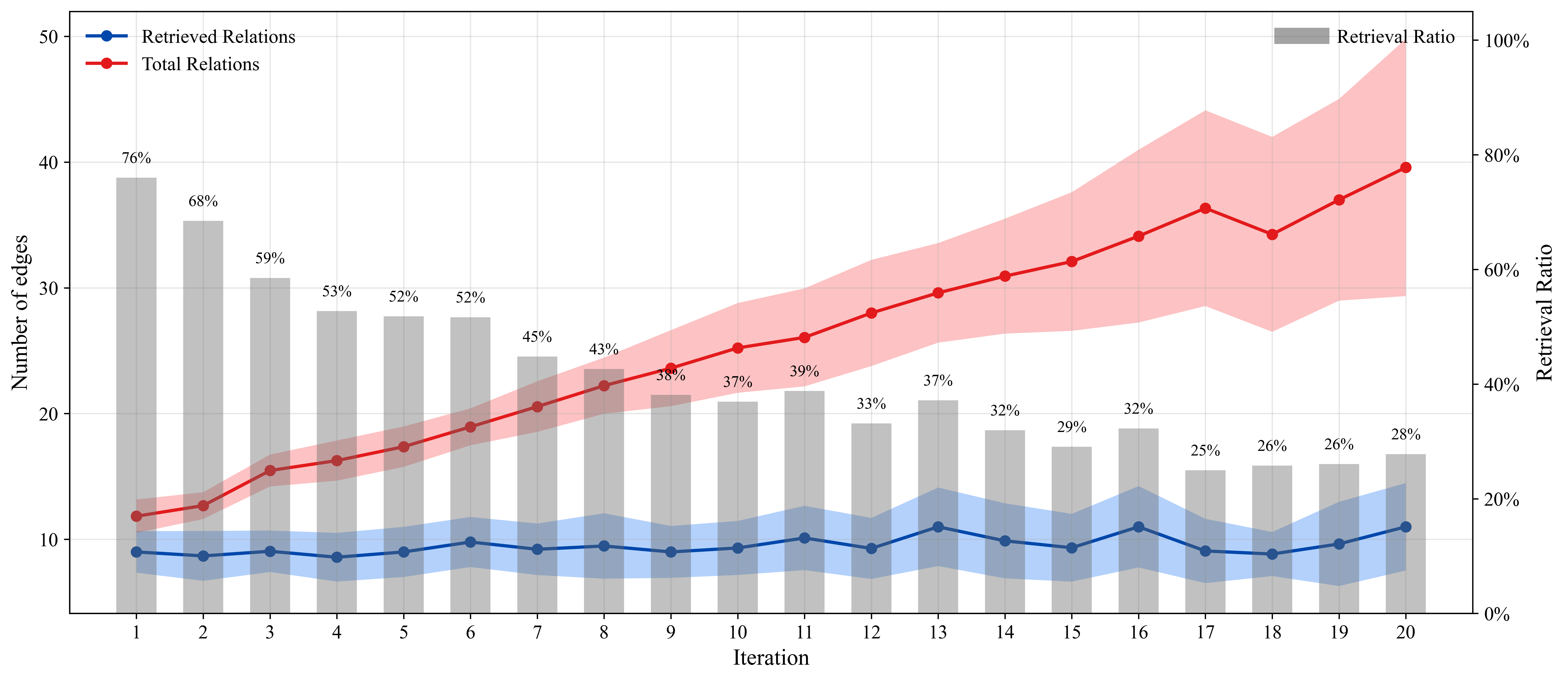}
    \caption{Average relationships related to update in spatial memory in each step.}
    \label{fig:spatial_analysis}
\end{figure*}

\subsection{Additional Efficiency analysis for Dynamic Spatial KG update algorithm}

We analyze the evolution of the spatial KG during long trajectories in EB-ALFRED, focusing on the first 20 iterations (with 95\% confidence intervals). As shown in Figure~\ref{fig:spatial_analysis}, the total number of spatial relationships in the KG (red line) increases gradually over iterations as RoboMemory is exploring the environment. In contrast, the number of relationships retrieved for update at each iteration (blue line) remains relatively stable, typically ranging around 10 edges per iteration. This stability is achieved because our method only updates a local subgraph relevant to the current observation.

We define the retrieval ratio as the proportion of relationships updated at each iteration relative to the total number of relationships in the KG. As shown in Figure~\ref{fig:spatial_analysis}, this ratio (illustrated by gray bars) decreases steadily from 76\% initially to 28\% at iteration 20. This trend indicates that, as the KG grows, each update affects a progressively smaller fraction of the entire graph. This demonstrates that our spatial KG update mechanism effectively localizes modifications, ensuring computational efficiency and mitigating interference through context-aware incremental updates.

\subsection{System-Level Inference Cost}

At each high-level decision, the Qwen-only, Planner--Critic, and full RoboMemory settings require 1.00, 1.16, and 9.72 VLM calls on average, corresponding to 4.0k, 4.81k, and 22.99k tokens, respectively. A forced-serial implementation of the memory pipeline is 2.1$\times$ slower than the parallel implementation. On the physical robot, high-level decision latency has a P50/P95 of 70/135 seconds, while the low-level VLA executor operates independently.

\subsection{Matched Attribution Controls}
\label{sec:matched_controls}

To further isolate the contributions of RoboMemory, we evaluate two matched controls on EB-ALFRED. The Planner--Critic-only control retains closed-loop planning without the RoboMemory memory system. For the Flat Spatial RAG control, we replace the dynamic Spatial KG with a vector database that stores the same spatial relations as independent entries and retrieves them using standard vector similarity, while keeping the remaining components unchanged.

\begin{table}[ht]
\centering
\caption{Matched attribution controls on EB-ALFRED (SR, \%).}
\label{tab:matched_controls}
\begin{tabular}{lccc}
\toprule
Condition & Avg. & Base & Long \\
\midrule
Planner--Critic, no RoboMemory & 38 & 62 & 14 \\
Flat Spatial RAG               & 48 & 56 & 40 \\
\textbf{RoboMemory (Dynamic KG)} & \textbf{67} & \textbf{68} & \textbf{66} \\
\bottomrule
\end{tabular}
\end{table}

The Planner--Critic-only control reaches 38\% average SR, while replacing the dynamic Spatial KG with Flat Spatial RAG reaches 48\%. In comparison, RoboMemory with the dynamic KG achieves 67\%, supporting the contributions of both structured memory and the incremental Spatial-KG design.

\subsection{Spatial Memory Diagnostics}
\label{sec:spatial_diagnostics}

To further evaluate the reliability of Spatial Memory updates, we directly test relation extraction and conflict resolution, the two key operations in our dynamic Spatial KG update pipeline.

Across 30 prompt-disjoint narratives containing 52 gold spatial relations, the relation extraction module achieves 90.7\% precision, 94.2\% recall, and 92.5\% F1. On 15 controlled stale, hallucinated, or mislabeled memory cases, the conflict resolver achieves a final F1 of 1.00 and Recovery@3 of 1.00, where Recovery@3 measures successful recovery within three attempts.

These results show that RoboMemory can reliably extract new spatial relations and correct inconsistent or outdated memory, supporting stable online maintenance of the dynamic Spatial KG.

\section{Dynamic Spatial Memory Update Algorithm}
\label{sec:kg}

\begin{algorithm*}
\caption{\textbf{Retrieval-based Incremental Knowledge Graph Update Algorithm}}
\begin{algorithmic}[1]
\REQUIRE New spatial knowledge graph \(G_{\text{new}} = (V_{\text{new}}, E_{\text{new}})\), main spatial knowledge graph \(G = (V, E)\), queries \(q \in Q\), entity \& query embeddings \(\mathcal{E}: V \cup Q \rightarrow \mathbb{R}^d\), maximum number of retrieved vertices \(n\), maximum k hops \(k\), VLM-based conflict resolver \(\text{ResolveConflict}(\cdot)\)
\ENSURE Updated consistent knowledge graph \(G'\)
\STATE $V_{\text{similar}} \leftarrow \bigcup_{q \in Q} \operatorname{TopK}_n \left( \{ v \in V \mid \text{cosine\_sim}(\mathcal{E}(q), \mathcal{E}(v)) \} \right)$
\COMMENT{For each query entity $q$, retrieve its top-$n$ most similar vertices in $G$ by cosine similarity of embeddings; take the union over all $q \in Q$.}
\STATE $V_{\text{expand}} \leftarrow \text{K-hop}_k(V_{\text{similar}}, G)$ 
\COMMENT{all nodes within $k$ hops from any node in $V_{\text{similar}}$}
\STATE \(V_{\text{retrieved}} \leftarrow V_{\text{similar}} \cup V_{\text{expand}}\)
\STATE $V_{\text{merged}} \leftarrow V_{\text{retrieved}} \cup V_{\text{new}}$
\STATE $G_{\text{union}} \leftarrow (V \cup V_{\text{new}},\, E \cup E_{\text{new}})$
\COMMENT{Combine the main graph and new observations into a unified graph.}
\STATE $G_{\text{local}} \leftarrow \text{InducedSubgraph}(V_{\text{merged}},\, G_{\text{union}})$
\COMMENT{Extract the subgraph induced by $V_{\text{merged}}$, containing all old and new edges among these nodes.}
\STATE $G_{\text{updated}} \leftarrow \text{ResolveConflict}(G_{\text{local}}, G_{\text{new}})$
\COMMENT{based on $G_{\text{new}}$, VLM update the relationship among different vertices in $G_{\text{local}}$}
\STATE $G' \leftarrow \left( G \setminus G_{\text{local}} \right) \cup G_{\text{updated}}$
\COMMENT{Replace the old subgraph in $G$ with the conflict-resolved updated subgraph.}
\STATE Remove isolated vertices from $G'$
\RETURN \(G'\)
\end{algorithmic}
\label{alg:KG-update}
\end{algorithm*}

Spatial Memory is a dynamically updated KG-based module designed to overcome agents' limitations in spatial reasoning. Specifically, our Spatial Memory is formulated as a directed KG \( G = (V, E) \), where \( V \) denotes the set of all objects in the environment. Each object is a vertex in the KG. Each object’s name is encoded into a single semantic embedding vector via a pretrained embedding model \citep{zhang2025qwen3}. The edge set \( E \) captures spatial relationships between objects, each represented as a triple (e.g., \([obj_1, \text{relationship}, obj_2]\)). To update the Spatial KG, we need to continuously extract new relationships from current observations and update the old relationships in the Spatial KG. We use a VLM-based conflict resolver to address this problem. However, the more relationships provided to the conflict resolver, the more time it needs to update the KG. So we need to update relationships that are only related to the current situation. We design an algorithm that retrieves a sub-graph of KG that includes all vertices related to the current situation and both old and new relationships among them. We provide the sub-graph and new relationships to the conflict resolver. We need the conflict resolver to update the sub-graph based on information from the new relationships. The algorithm is shown in Algorithm~\ref{alg:KG-update}.

To update \(G\), we make use of the information provided by the step summarizer and query generator from the information preprocessor introduced in Section~\ref{sec:information_preprocessor}. First, a pretrained VLM-based Relation Retriever extracts the latest spatial relationships \( G_{\text{new}} = (V_{\text{new}}, E_{\text{new}}) \) from the information provided by the step summarizer, which records high-level information in the current observation. Next, natural language queries (represented as \(q \in Q\)) (provided by the query generator) are used to retrieve relevant object vertices from \( G \) via cosine similarity search. We select the top \(n\) similar vertices compared with the query. These vertices are denoted as \(V_{\text{similar}}\).

Spatial KG maintains the relationships among different objects, so if we want to retrieve spatial information from spatial KG, we need to search for other objects that are related to the objects we observed in the current observation. In this way, we not only remember objects we can see, but also know the spatial information of the objects we cannot see. So we choose the k-hop algorithm to expand \(V_{\text{similar}}\) using a K-hop neighborhood algorithm to capture contextually related objects. The K-hop algorithm is represented as \(\text{K-hop}_k(V, G)\), which returns all vertices reachable within \(\leq k\) hops from any vertices in \(V\). The retrieved vertices are \(V_{\text{expand}}\). We combine the vertices retrieved by cosine similarity and their K-hop neighbors to \(V_{\text{retrieved}}\).

However, we need to resolve the conflict between new and old relationships. So \(V_{\text{retrieved}}\) and the relationships among \(V_{\text{retrieved}}\) is not enough. We need new vertices and relationships involved in the graph we provided to the VLM-based conflict resolver. To extract all vertices and relationships for the VLM-based conflict resolver and relationships, we not only need the relationships from KG (old information) and \(G_{\text{new}}\), which represent new information. We need to connect old information and new information. To achieve this goal, we merge \(G_{\text{new}}\) to \(G\), which add new edges and vertices to \(G\). We denote the merged KG as \(G_{\text{union}}\). In \(G_{\text{union}}\), we mix out-of-date and latest information. Then, we extract an induced sub-graph of \(V_{\text{merged}} = V_{\text{retrieved}} \cup V_{\text{new}}\) from \(G_{\text{union}}\). As both vertices from old graph \(G\) and new graph is mixed in \(V_{\text{merged}}\) and both edges from old and new graph is in \(G_{\text{union}}\), the retrieved induced sub-graph \(G_{\text{local}}\) contains all out-of-date and latest relationships among vertices that is related to current situation.

As \(G_{\text{local}}\) contains all out-of-date and latest relationships and those relationships may have conflicts, a VLM-based conflict resolver (represent as \(\text{ResolveConflict}(\cdot)\)) is designed to resolve conflicts in \(G_{\text{local}}\), and make sure that the relationship is the latest. The conflict resolver will take in \( G_{\text{new}} \) and \( G_{\text{local}} \), where \( G_{\text{local}} \) is the graph waiting for update and \( G_{\text{new}} \) provide update signal. The VLM-based conflict resolver will perform necessary updates such as adding vertices or inserting, deleting, or modifying relationships in \( G_{\text{local}} \) based on \( G_{\text{new}} \). The reconciled subgraph is then merged back into \( G \), and any vertices that have lost all connections to other vertices during the update are pruned.

This design offers two key advantages: (1) Efficiency via localized updates: By restricting modifications to a context-relevant subgraph, we significantly reduce the number of relationships processed per update. Since VLMs struggle with reasoning over large sets of relationships, this constraint substantially improves both the efficiency and effectiveness of VLM-based KG updates. (2) Dynamic adaptability: The system continuously maintains up-to-date spatial knowledge, enabling agents to operate robustly in dynamic real-world environments.

\begin{figure*}[ht]
    \centering
    \includegraphics[width=1.0\linewidth]{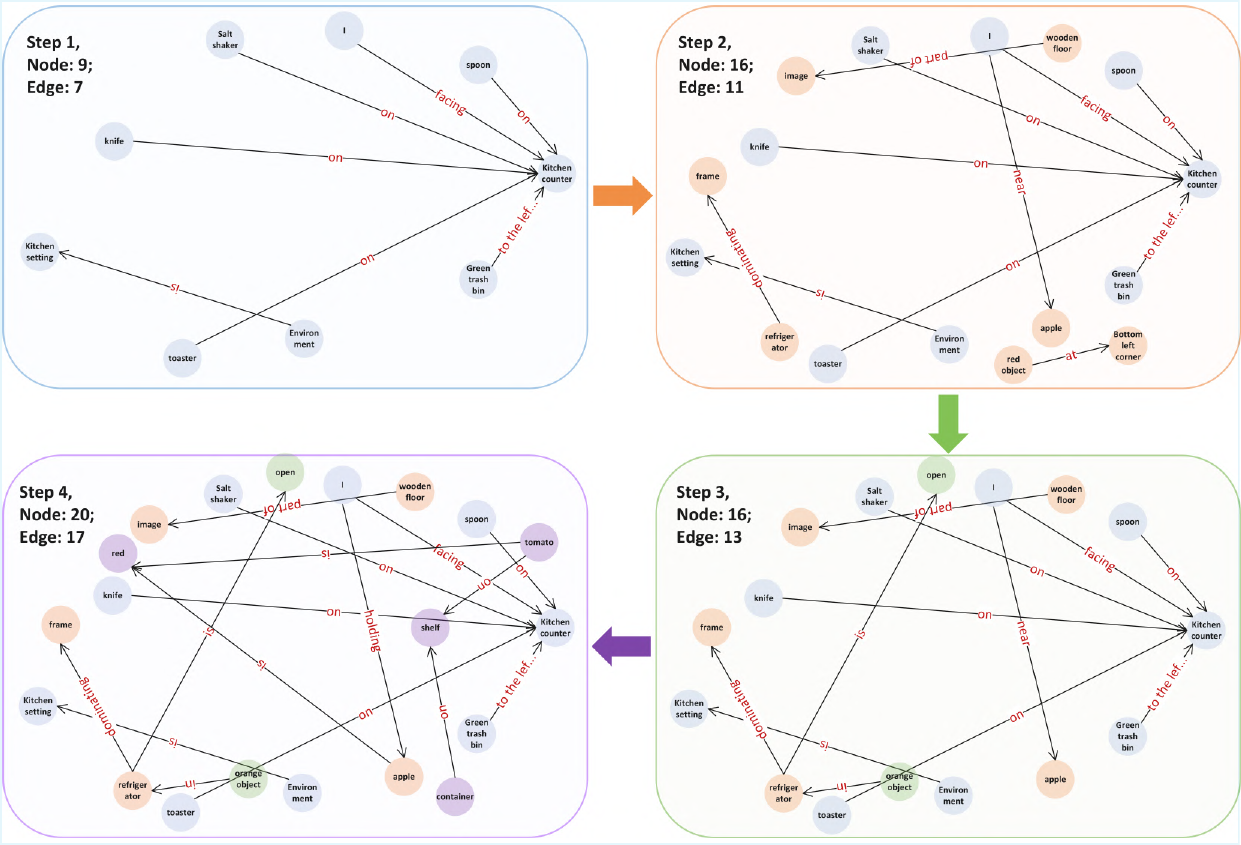}
    \caption{Visualization of Spatial Memory's dynamic update process.}
    \label{fig:kg-example}
\end{figure*}

\subsection{Example of Dynamic Spatial Memory Update process}

In RoboMemory's Spatial Memory, the KG is dynamically constructed during environment exploration. As illustrated in Figure~\ref{fig:kg-example}, we demonstrate the progressive expansion of the KG in Spatial Memory as the agent navigates through the environment. The figure indicates a continuous growth in the number of both vertices and edges of the KG as exploration progresses.

Notably, the KG undergoes dynamic updates through RoboMemory's environmental interactions. For example, the initial KG state displays the relation ``I am near the apple. But as the agent picks up the apple in the third step, in the fourth KG, the relationship becomes ``I hold the apple". This demonstrates RoboMemory's capability for dynamic KG maintenance and expansion.

By querying this KG, the Planner-Critic module gains access to rich spatial information, empowering RoboMemory with robust spatial memory capabilities that significantly enhance its performance in EmbodiedBench environments.

\section{Proof of Dynamic Spatial Memory Update Algorithm}
\label{sec:proof}

\begin{theorem}[Upper Bound on K-hop Vertex Extraction in Directed Graphs]
Let \( G = (V, E) \) be a finite directed graph with maximum out-degree \( D \geq 1 \), and let \( \mathcal{S} \subseteq V \) be a set of \( M \) source vertices. Define the \emph{K-hop neighborhood} \( \mathcal{N}_K(s) \) of a vertex \( s \in \mathcal{S} \) as the set of vertices reachable from \( s \) via directed paths of length at most \( K \). Then the total number of distinct vertices in the union of all K-hop neighborhoods,
\[
\mathcal{N}_K(\mathcal{S}) = \bigcup_{s \in \mathcal{S}} \mathcal{N}_K(s),
\]
Satisfies the following upper bound:
\[
|\mathcal{N}_K(\mathcal{S})| \leq 
\begin{cases}
M \cdot \dfrac{D^{K+1} - 1}{D - 1}, & \text{if } D > 1, \\
M \cdot (K + 1), & \text{if } D = 1.
\end{cases}
\]
\end{theorem}

\begin{proof}
For any vertex \( s \in \mathcal{S} \), the number of distinct vertices reachable from \( s \) within \( i \) hops is at most \( D^i \), assuming the worst-case scenario where each vertex encountered has the maximum out-degree \( D \), and all neighbors are distinct and non-overlapping.

Thus, the size of the K-hop neighborhood of a single vertex satisfies:
\[
|\mathcal{N}_K(s)| \leq \sum_{i=0}^{K} D^i = 
\begin{cases}
\dfrac{D^{K+1} - 1}{D - 1}, & \text{if } D > 1, \\
K + 1, & \text{if } D = 1.
\end{cases}
\]

Since there are \( M \) such source vertices and assuming no overlaps between their K-hop neighborhoods (worst case), the union size satisfies:
\[
|\mathcal{N}_K(\mathcal{S})| \leq M \cdot |\mathcal{N}_K(s)|.
\]

Substituting the bound on \( |\mathcal{N}_K(s)| \) gives the result.
\end{proof}
\vspace{0.5cm}

\begin{theorem}[Upper Bound for K-hop Vertex Extraction in Normalized Directed Graphs]
Let \( G = (V, E) \) be a finite directed graph with \( |V| = n \) vertices. Assume the maximum out-degree is at most \( D_{\max} = Dn \), and the maximum in-degree is at most \( N_{\max} = Nn \), where \( D, N \in (0, 1] \) are constants. Let \( \mathcal{S} \subseteq V \) be a set of \( M \) source vertices. Define \( \mathcal{N}_K(\mathcal{S}) \) as the union of all vertices reachable from \( \mathcal{S} \) via paths of length at most \( K \), using only outgoing edges. Then the number of extracted vertices satisfies:
\[
|\mathcal{N}_K(\mathcal{S})| \leq \min \left\{ n,\; M \cdot \frac{(Dn)^{K+1} - 1}{Dn - 1} \right\}.
\]
In particular, when \( Dn \gg 1 \), we have the approximation:
\[
|\mathcal{N}_K(\mathcal{S})| \lessapprox M \cdot (Dn)^K.
\]
\end{theorem}

\begin{proof}
For each vertex \( s \in \mathcal{S} \), the maximum number of reachable vertices within \( i \)-hops is at most \( (Dn)^i \) under the assumption of maximum out-degree and no overlap.

Summing over hops from 0 to \( K \), we get for each root:
\[
|\mathcal{N}_K(s)| \leq \sum_{i=0}^{K} (Dn)^i = \frac{(Dn)^{K+1} - 1}{Dn - 1}.
\]

Assuming no overlap among the \( M \) source vertex expansions (worst case), we have:
\[
|\mathcal{N}_K(\mathcal{S})| \leq M \cdot \frac{(Dn)^{K+1} - 1}{Dn - 1}.
\]

Since the total number of vertices in the graph is \( n \), this quantity is also trivially bounded above by \( n \), yielding the result.
\end{proof}

\section{Additional Environment settings}
\label{sec:add_env_set}

\subsection{EB-ALFRED and EB-Habitat}

We adopt the same environment parameters as in EmbodiedBench. The maximum steps per task are set to 30, with image inputs of size 500 \(\times\) 500. The temporal memory buffer length is set to 4. In addition, to align the simulation setting with the function-call interface used in real-world deployment, the global action space is decomposed into action types and target objects. Each action is represented as a Python-style function that takes an object as its argument. Candidate target objects are provided to the Agent, which selects the appropriate action and target object based on the task requirement. This design provides a consistent action interface across simulated and real-world environments. The detailed action APIs are presented in Table~\ref{tab:action_commands}.

Since EB-ALFRED and EB-Habitat provide comprehensive high-level action APIs, we do not employ the VLA-Based Low-Level Executor in these environments. Instead, we utilize the built-in low-level controllers from EmbodiedBench.

\subsection{Baselines} 

To facilitate comparisons, we consider two types of baselines. First, we choose the advanced closed-source and open-source VLMs as a single agent. We compare their performance with RoboMemory. For closed source VLMs, we choose GPT-4o and GPT-4o-mini \citep{GPT-4o-mini,hurst2024gpt}, Claude3.5-Sonnet and Claude-3.7-Sonnet \citep{Claude-3.5-Sonnet}, Gemini-1.5-Pro and Gemini-2.0-flash \citep{team2024gemini,Gemini2.0}. For open source VLMs, we choose Llama-3.2-90B-Vision-Ins \citep{llama3.2}, InternVL-2.5-78B/38B \citep{chen2024expanding}, InternVL-3-78B \citep{zhu2025internvl3}, and Qwen2.5-VL-72B-Ins \citep{bai2025qwen2}. Secondly, we choose four agent frameworks: (1) Reflexion \citep{shinn2024reflexion}, which introduces a simple long-term memory and a self-reflection module. Reflexion uses the self-reflection module to summarize experiences as long-term memory, thereby enhancing the model's capabilities. (2) Voyager \citep{wang2023voyager}, which utilizes a skill library as its procedural memory, is a widely used baseline for embodied agent planning. (3) Cradle \citep{tan2024cradle}, which proposes a general agent framework with episodic and procedural memory and gains good performance at various multi-modal agent tasks. (4) RoboOS \citep{tan2025roboos}, which proposes an embodied agent framework that consists of a Hierarchy Scene-Graph based Spatial Memory.

In our experiments, each agent framework is tested using Qwen2.5-VL-72B-Ins  with temperature set as 0. For the RoboOS framework, we test it on RoboBrain2-32B \citep{team2025robobrain}, where the RoboBrain2-32B model is designed for the RoboOS framework.

The Qwen2.5-VL-72B-Ins represents a high-performing open-source alternative. Notably, the Qwen2.5-VL-72B-Ins demonstrates performance comparable to advanced closed-source VLMs in several benchmark tasks \citep{white2024livebench}. We use the Qwen3-Embedding model \citep{zhang2025qwen3} to create embedding vectors for RAGs in RoboMemory. For the Low-level Executor, since EB-ALFRED provides high-level action APIs, we use the Low-level Executor provided by EmbodiedBench instead of the VLA-based method. 

We use the standard metrics from EmbodiedBench. More detail is provided in Appendix~\ref{sec:metrics}

\subsection{Metrics}
\label{sec:metrics}

We define two evaluation metrics to assess the performance: (1) Success Rate (SR), which is the ratio of completed tasks to the total number of tasks in each difficulty level. This metric reflects the agent's ability to complete tasks across randomly generated scenarios. (2) Goal Condition Success Rate (GC), which is the ratio of intermediate conditions achieved to the maximum possible score in each scenario. A GC of 100\% indicates that the task is completed in the given scenario. These two metrics can be computed as:

\begin{equation}
    SR = \mathbb{E}_{x \in \mathcal{X}} \left[ \mathds{1}_{SCN_x = GCN_x}\right]
\end{equation}

\begin{equation}
    GC = \mathbb{E}_{x \in \mathcal{X}} \left[ \frac{SCN_x}{GCN_x}\right]
\end{equation}

\noindent In above formulas, \(\mathcal{X}\) denotes the test subset, and \(x\) represents a test task. The success condition number (\(SCN_x\)) refers to the number of conditions the agent has accomplished, while the global condition number (\(GCN_x\)) indicates the total number of conditions required for task completion. The task is considered successful if \(SCN_x = GCN_x\). 

\subsection{Additional Settings for Baselines}

\textbf{EB-ALFRED and EB-Habitat.} For our single VLM-Agent baseline, we utilized the reported results from the EmbodiedBench paper to establish a consistent benchmark, where the agent relies on a basic interaction history as its memory module. For other VLM frameworks, we replicated the experimental setups as described in both EmbodiedBench and the respective original papers. Crucially, to familiarize all baseline agents with the EmbodiedBench environment, we supplemented them with a few-shot example and a comprehensive catalog of actionable objects—applying the exact same conditions as those used for the single VLM-Agent benchmark.

For VLM-Agent Frameworks, we use the same VLMs (listed in Table~\ref{tab:combined_full_benchmarks}) to power all modules that require VLM control.

\subsection{Real-world experiments}

\begin{table*}[ht]
\centering
\caption{Robot Action Command For different environments}
\label{tab:action_commands}
\begin{tabular}{|l|c|c|c|}
\hline
\textbf{Action Type} & \textbf{EB-ALFRED} & \textbf{EB-Habitat} & \textbf{Real World} \\
\hline
Navigation & \texttt{find(obj)} & \texttt{navigate(point)} & \texttt{navigate\_to(point)} \\
Pick Up Object & \texttt{pick\_up(obj)} & \texttt{pick(obj)} & \texttt{pick\_up(obj)} \\
Drop to Ground & \texttt{drop()} & -- & -- \\
Place to Receptacle & \texttt{put\_down()} & \texttt{place(rec)} & \texttt{put\_down\_to(rec)} \\
Open Object & \texttt{open(obj)} & \texttt{open(obj)} & \texttt{open(obj)} \\
Close Object & \texttt{close(obj)} & \texttt{close(obj)} & \texttt{close(obj)} \\
Turn On & \texttt{turn\_on(obj)} & -- & \texttt{turn\_on(obj)} \\
Turn Off & \texttt{turn\_off(obj)} & -- & \texttt{turn\_off(obj)} \\
Slice Object & \texttt{slice(obj)} & -- & -- \\
Task Complete & -- & -- & \texttt{task\_complete()} \\
\bottomrule
\end{tabular}
\end{table*}

We construct a common kitchen scenario to evaluate how RoboMemory reuses accumulated experience to support interactive environmental learning in real-world settings. The environment contains 5 navigable points, 8 interactive objects, and over 10 non-interactive (but potentially distracting) items.  We created three task categories (5 tasks each). Tasks are matched to EB-ALFRED’s Base subset (avg. oracle: 10–20 steps), though actual executions often exceed 20 steps due to search and error recovery. In the real world, we use interactive environmental video recordings captured during action execution (rather than static snapshots taken after action completion) as RoboMemory's input. This provides a more temporally coherent perception. 

We use Mobile ALOHA \citep{fu2024mobile} as our physical robotic platform, we design three categories of tasks: (1) Pick up \& put down: The agent must locate a specified object among all possible positions and place it at a designated location. This task tests the model's basic object-searching and planning abilities. (2) Pick up, operate \& put down: Building upon the first task, the agent must additionally perform operations such as heating or cleaning the object. This task requires longer-term planning, which is crucial in embodied environments. (3) Pick up, gather \& put down: The agent must place specified objects into a movable container and then move the container to a target location. This task evaluates the agent's understanding of object relationships, requiring it to remember the positions of at least two objects (the container and the target item) and their spatial relationship. For each type of task, we design 5 tasks. So our experiments include 15 long-term real-world tasks.

To adapt to the real-world setup, we define high-level action APIs similar to those in EB-ALFRED and EB-Habitat. Additionally, we train a VLA-based model to execute tasks according to our action APIs. The detailed action APIs are presented in Table~\ref{tab:action_commands}.

For the low-level executor, we use one main camera and two arm-mounted cameras as input, each with a resolution of 640 \(\times\) 480. The temporal memory buffer length is set to 4. All real-world settings use the same low-level executor, navigation controller, action interface, and maximum step budget.

In our experiments, we set the maximum steps per task to 25. We also provide an API for actively terminating tasks. Since real-world environments lack direct success/failure feedback, RoboMemory must autonomously determine task completion. To prevent excessively long task execution, we enforce termination after 25 steps if no success is achieved. A single main camera (640 \(\times\) 480 resolution) records video during action execution as input for RoboMemory's higher-level processing. The reported task success is evaluated independently based on the final environment state, following the state-based success criteria used in EB-ALFRED.

\subsection{Training Details of Low-Level Executor}

We use the $\pi_0$ model as our foundation model. We collected 1,040 data samples over 10 types of tasks for fine-tuning. We use LoRA fine-tuning to save resources during fine-tuning. The specific fine-tuning parameters and action types are given in Table~\ref{tab:dataset_hyperparams}. For tasks involving both pick-up and place actions, we split these tasks into separate pick-up and place actions. These are then treated as two distinct data samples during training. The separation of pick-up and place action allows the VLA to carry an object in its hand. For training, we used a server with six A100-80GB GPUs. The total training time was 12 hours.

Besides, we use the built-in LiDAR SLAM system of the Mobile ALOHA robot base as the navigation action actuator. We define five typical navigation points, similar to EB-Habitat. We used SLAM to navigate between these navigation points.

\begin{table*}[!htb] 
\caption{Dataset statistics and training hyperparameters for robotic manipulation tasks.} 
\label{tab:dataset_hyperparams} 
\centering 
\renewcommand{\arraystretch}{1.2} 
\begin{tabular}{|l|r|l|l|} 
\hline 
\multicolumn{2}{|c|}{\textbf{Dataset Statistics}} & \multicolumn{2}{c|}{\textbf{Training Configuration}} \\ 
\hline 
\textbf{Action Type} & \textbf{\#Episodes} & \textbf{Parameter} & \textbf{Value} \\ 
\hline 
Turning on/off faucet & 142 & Optimizer & AdamW \\ 
Picking up \& Placing basket on counter & 63 & Batch size & $32 \times 6$ \\ 
Picking up \& Placing basket in sink & 72 & Training steps & 10,000 \\ 
Picking up \& Placing banana into basket & 114 & Learning rate & $6.12 \times 10^{-5}$ \\ 
Throwing bottle into trash bin & 132 & warm up step & 500\\ 
\cline{3-4} 
Placing gum box on dish & 120 & \multicolumn{2}{c|}{\textbf{LoRA Configuration}} \\ 
\cline{3-4} 
Picking up \& Placing cup on plate & 51 & rank & 16 \\ 
Picking up \& Placing dish into sink & 69 & $\alpha$ & 16\\ 
\cline{3-4} 
Throwing paper ball into trash bin & 135 & \multicolumn{2}{c|}{\textbf{Resource Usage}} \\ 
\cline{3-4} 
Open/close oven & 142 & GPU & A100-80GB $\times$ 6 \\ 
\cline{1-2}
\textbf{Total episodes} & \textbf{1040} & Training time & 12 hours \\ 
\hline 
\end{tabular} 
\vspace{-0.5em} 
\end{table*}

\subsection{Hyperparameters of RoboMemory}








In this section, we describe the hyperparameter settings for the upper brain of RoboMemory: the information preprocessor and the Comprehensive Embodied Memory. Importantly, we use a unified set of hyperparameters across all experimental settings, including EB-ALFRED, EB-Habitat, and real-world deployments.

The information preprocessor consists of two components: a step summarizer and a query generator. Given multimodal inputs at each step, the step summarizer produces a single natural language description, while the query generator concurrently formulates \(4 - 5\) distinct natural language queries.

The Comprehensive Embodied Memory integrates four memory modules: Temporal, Spatial, Semantic, and Episodic memory. Temporal memory is implemented as a fixed-size buffer with a maximum capacity of \(4\) entries. For Spatial Memory, during similarity-based retrieval, we first identify the top \(N = 3\) most relevant vertices and then perform a K-hop graph traversal with \(K = 2\). The Episodic memory retrieves the top \(N = 5\) most relevant past experiences for each query. The Semantic memory maintains hierarchical summaries at both the action and task levels; during retrieval, it returns \(N_s = 2\) action-level and \(N_t = 2\) task-level summaries. Furthermore, memory updates (e.g., insertion, modification, or deletion) are applied only to the top \(N_{\text{update}} = 10\) most relevant entries in the Semantic memory to ensure efficiency and coherence.

\section{Supplementary Examples for Qualitative Analysis}
\label{sec:case}

\subsection{Real World}

\begin{figure*}[ht]
    \centering
    \includegraphics[width=0.9\linewidth]{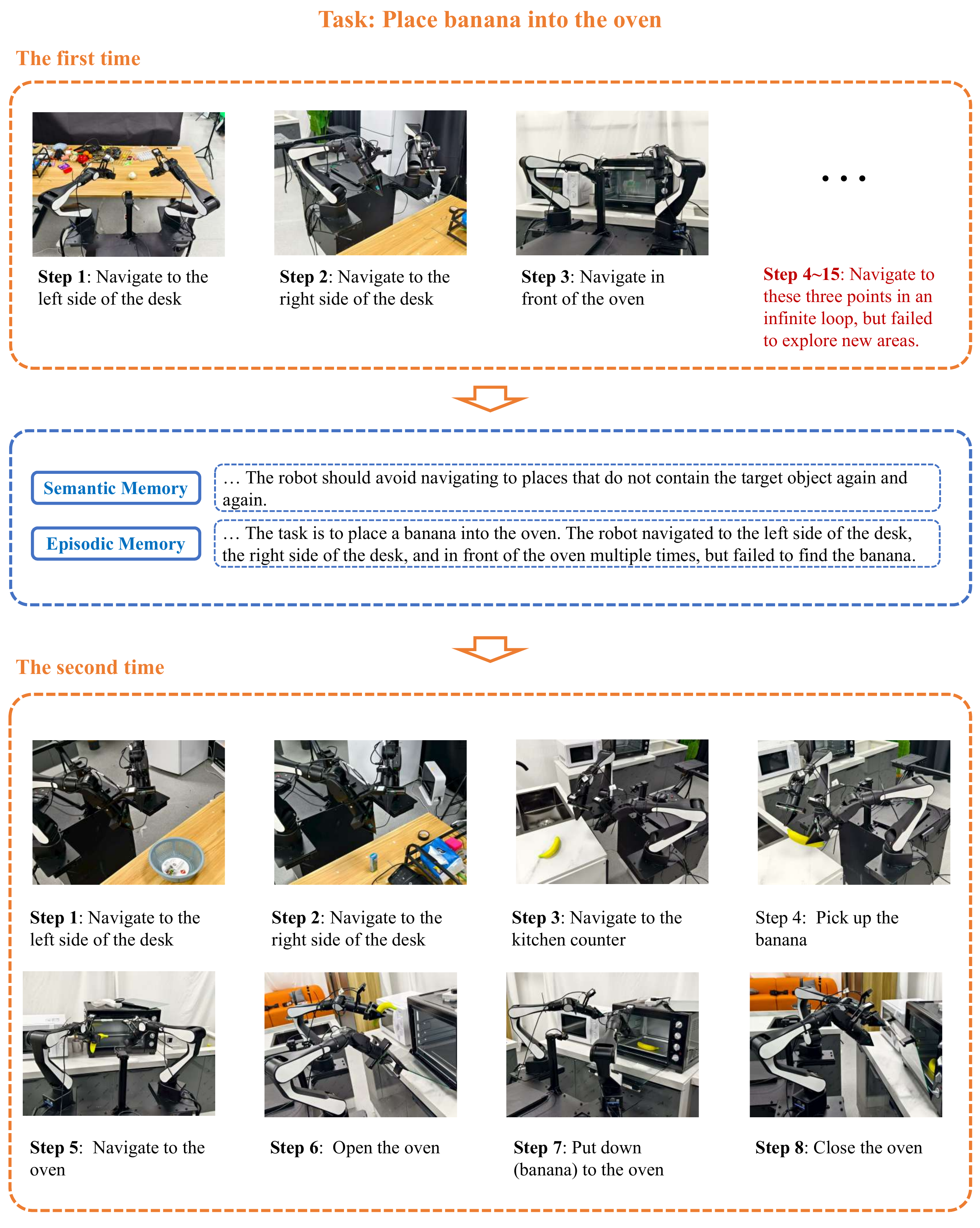}
    \caption{Case that a task is failed, but the experience can help RoboMemory to succeed in the next try.}
    \label{fig:case-1}
\end{figure*}

In Figure~\ref{fig:case-1}, we demonstrate an example of RoboMemory learning through trial and error in a real-world environment. Our task is ``place a banana into the oven." This task required RoboMemory to complete the objectives of finding the banana, picking it up, and transporting it to the oven. We observed that RoboMemory became stuck in an infinite loop during the first attempt. The banana was randomly placed on the ``kitchen counter," but RoboMemory overlooked this navigation target and remained trapped, exploring other navigation targets instead.

However, based on this bad attempt, the semantic memory summarized that the robot should not repeatedly search in locations where the ``banana" could not be found. Meanwhile, the episodic memory recorded what RoboMemory had done and the outcomes during the first attempt. Based on the information provided by semantic and episodic memory, in the second attempt, RoboMemory recognized that it had not previously tried navigating to the ``kitchen counter." After attempting this, it successfully completed the task. This example illustrates the role of RoboMemory's long-term memory.

\begin{figure*}[ht]
    \centering
    \includegraphics[width=0.9\linewidth]{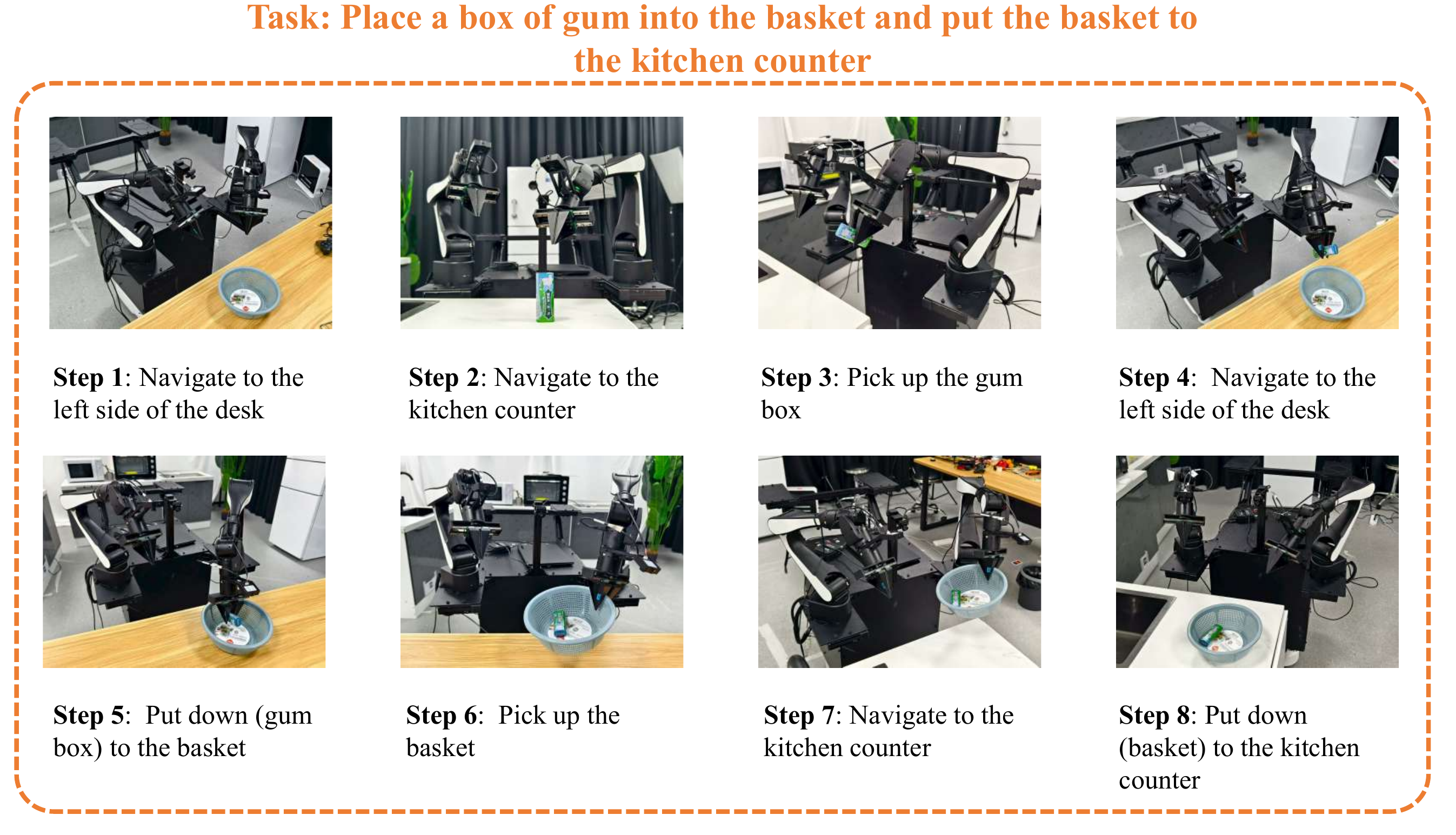}
    \caption{Case that a task is successful.}
    \label{fig:case-2}
\end{figure*}

We also provide an example that completes the task in the first attempt. The example is shown in Figure~\ref{fig:case-2}. This example demonstrates that the RoboMemory has the ability to handle some relatively complex tasks in the real world. The task in this example is ``Place a box of gum into the basket and put the basket on the kitchen counter". Because two objects in different positions are involved in this task, RoboMemory has to memorize the position of at least one object to achieve the goal. With the help of the spatial memory, RoboMemory completes the task successfully.

\subsection{EB-ALFRED}

We select three examples in EB-ALFRED to show the errors that RoboMemory may encounter and the reasons why or why not RoboMemory can achieve the goal.

\begin{figure*}[ht]
    \centering
    \includegraphics[width=0.9\linewidth]{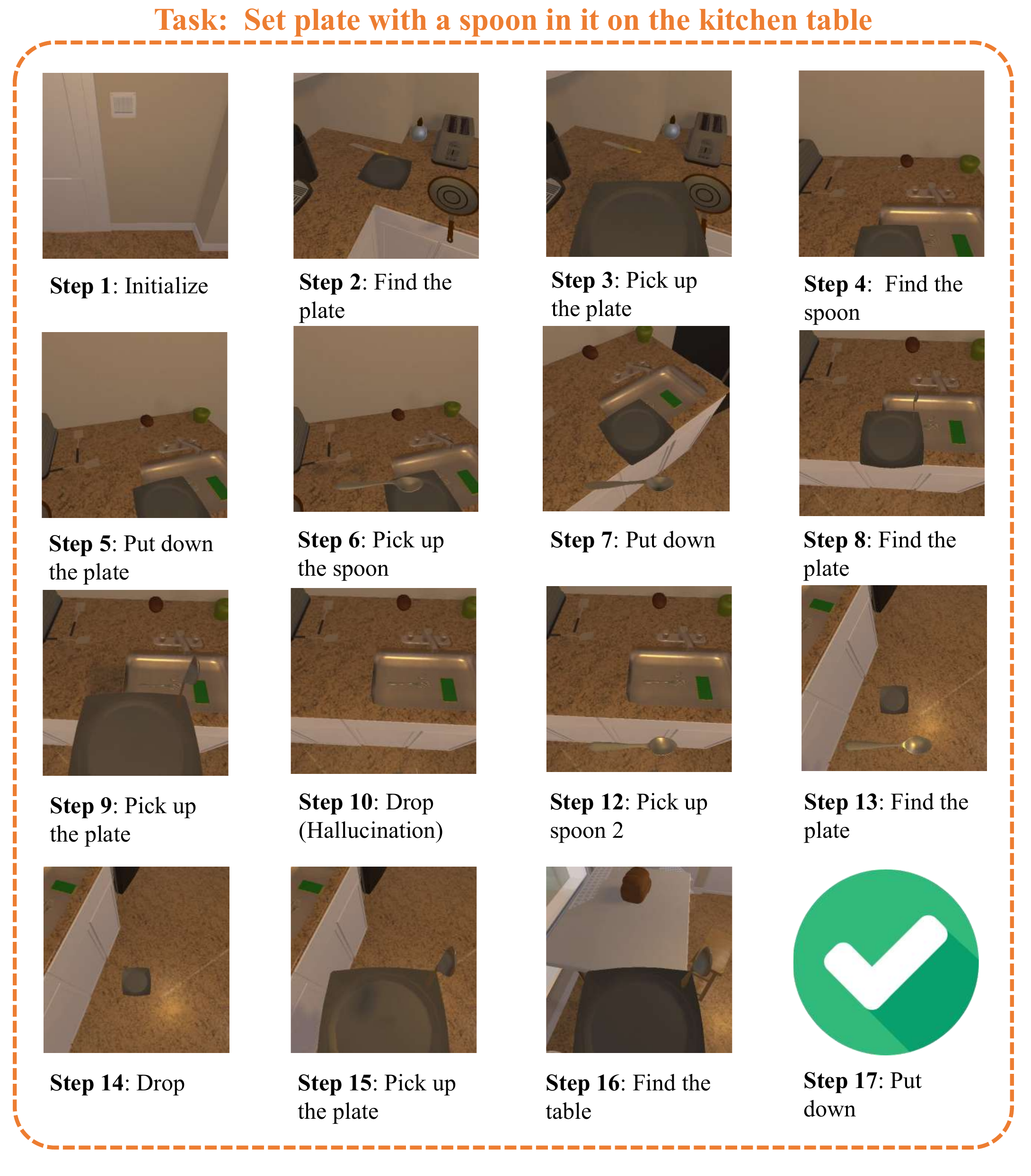}
    \caption{Case that a task is successful with the help of the critic and spatial memory modules.}
    \label{fig:case-4}
\end{figure*}

\noindent \textbf{Successful example.} We select a successful example to show how RoboMemory performed in the EB-ALFRED environment. The example trajectory is shown in Figure~\ref{fig:case-4}. 

The task of this example is ``set a plate with a spoon on it on the kitchen table". However, in step 10, the Planner seems to ignore the temporal information from memory modules. RoboMemory thinks that it still needs to pick up the spoon (even though it has already placed a spoon in the plate). However, with the help of the critic, it finally becomes aware that picking up another spoon is redundant, so RoboMemory goes back to the current trajectory and successfully completes the task at the end. 

In this example, RoboMemory successfully overcame the hallucination and eventually achieved the goal. This example demonstrates that the critic module can help RoboMemory to overcome error cases.

\begin{figure*}[ht]
    \centering
    \includegraphics[width=0.9\linewidth]{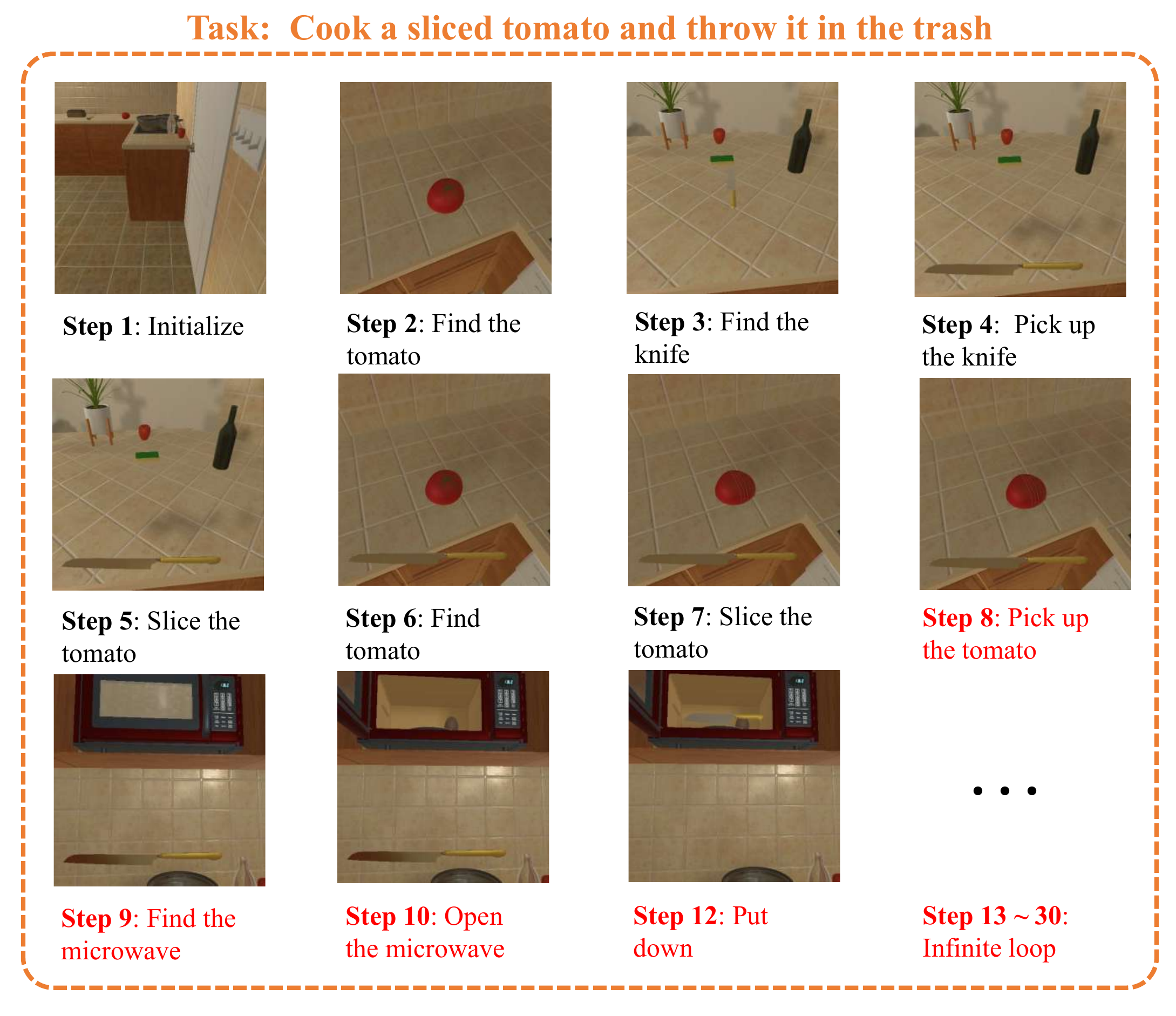}
    \caption{Case that a task fails in an infinite loop because the critic module failed to stop the agent when its planned action is no longer suitable.}
    \label{fig:case-3}
\end{figure*}

\noindent \textbf{Failed example.} We demonstrate a representative example of the Critic Error. The example trajectory is shown in Figure~\ref{fig:case-3}. In this example, the task involves slicing and heating a tomato and moving the heated tomato slice to the trash can. Initially, RoboMemory successfully sliced the tomato with a knife. But when the planner plans the whole sequence, it forgets to drop the knife before picking up the tomato (this is necessary because in EB-ALFRED, the robot can only hold one object at a time). The critic and the planner should notice this situation and ask the planner to replan, as RoboMemory failed to pick up a tomato slice. However, the critic module ignores this issue, and thus, after it heats the knife instead of a tomato slice, it gets stuck in an infinite loop.

\begin{figure*}[ht]
    \centering
    \includegraphics[width=0.9\linewidth]{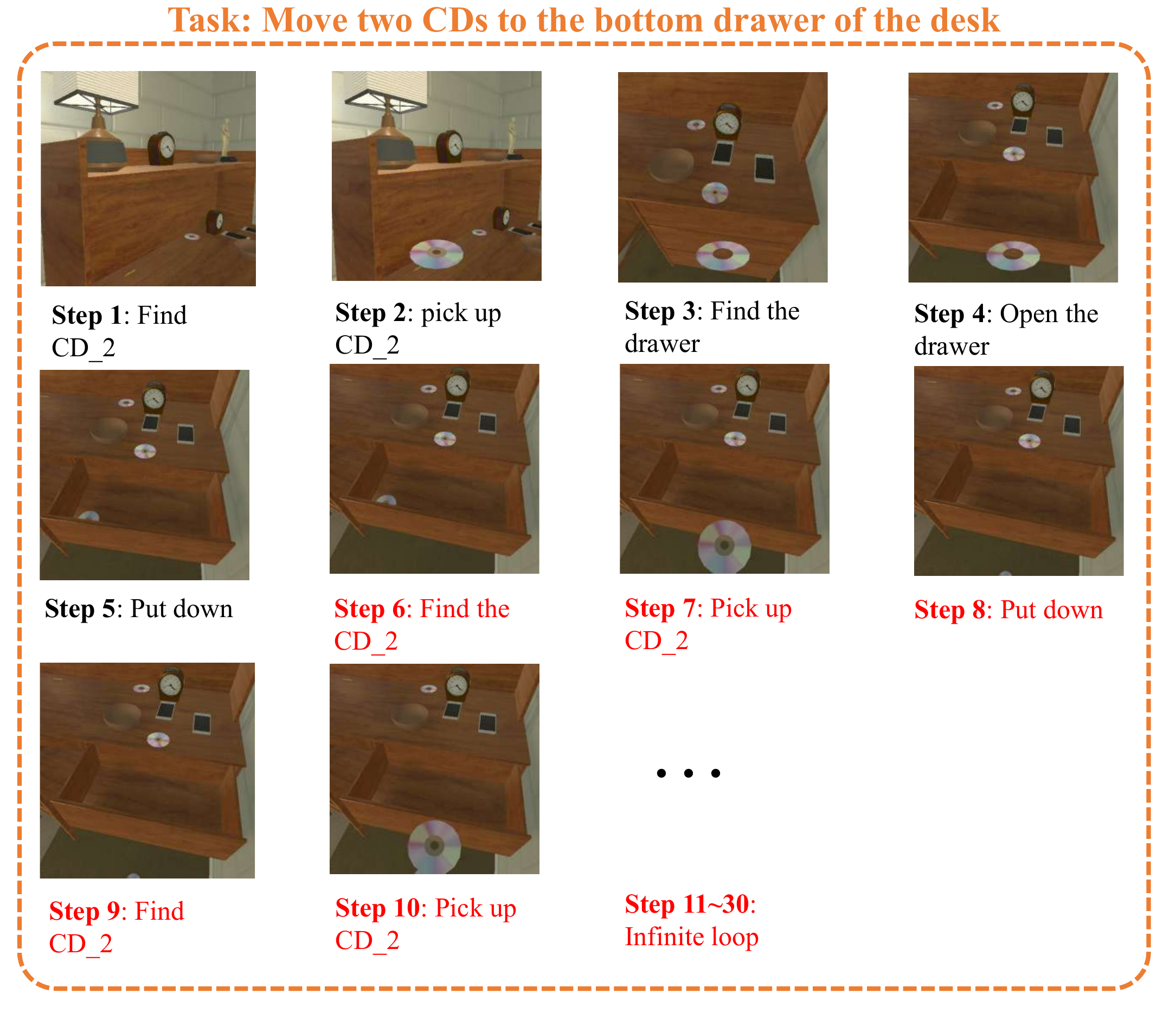}
    \caption{Case that a task fails in an infinite loop because of inaccurate action planning.}
    \label{fig:case-5}
\end{figure*}

Besides, we provide another example demonstrating a representative failure caused by inaccurate action planning. The example trajectory is shown in Figure~\ref{fig:case-5}. In the trajectory, RoboMemory is asked to place two CDs into the drawer. However, at step 6, the robot failed to select correct CD object. In this experiment, RoboMemory has already put CD\_2 into the drawer, but it keeps picking up  CD\_2 even though the memory has clearly indicated that CD\_2 has already been put down. So we classify this as inaccurate action error. This indicates that the planner failed to comprehensively integrate information from both the memory and information-gathering modules, resulting in inaccurate action planning.

\end{document}